\documentclass[12pt]{article}
\usepackage[usenames,dvipsnames]{color}

\usepackage{graphicx}
\usepackage{amsthm}
\usepackage{graphics}
\usepackage{amsmath,,amssymb}
\usepackage{algorithmic}
\usepackage{algorithm}
\usepackage{caption}
\usepackage{subcaption}
\usepackage{pstricks, pst-node}
\usepackage{tikz}
\usepackage[linewidth=1pt]{mdframed}
\usepackage{multirow}
\usepackage[round,colon,authoryear]{natbib}
\usepackage{pgfplots}
\usepackage{hyperref}
\usepackage{bm}
\usepackage{enumitem}
\usepackage[title]{appendix}

\pgfplotsset{compat=1.10}
\usepgfplotslibrary{fillbetween}
\usetikzlibrary{patterns}
\usetikzlibrary{shapes,arrows}
\usetikzlibrary{arrows.meta}

\newtheorem{theorem}{Theorem}
\newtheorem{proposition}{Proposition}

\newtheorem{assumption}{Assumption}

\def\EE{\mathbb{E}}
\def\PP{\mathbb{P}}
\def\RR{\mathbb{R}}
\def\Rcal{\mathcal R}

\def\Var{{\rm Var}}
\def\Cov{{\rm Cov}}

\def\Bcal{\mathcal B}

\def\Lcal{\mathcal L}
\def\Ical{\mathcal I}

\def\HH{\mathbb H}

\def\Mcal{\mathcal M}
\def\Pcal{\mathcal P}

\def\Scal{\mathcal S}
\def\Ncal{\mathcal N}
\def\Tcal{\mathcal T}

\def\Hcal{\mathcal H}

\def\bI{\mathbf I}

\tikzstyle{block} = [rectangle, draw, fill=white!80!black, line width=2pt,
    text width=15em, text centered, rounded corners, minimum height=3em]
\tikzstyle{line} = [draw, -latex',line width=2pt]

\begin{document}
\title{On Linear Separation Capacity of Self-Supervised Representation Learning}
\author{Shulei Wang\\ University of Illinois at Urbana-Champaign}
\date{(\today)}

\maketitle

\footnotetext[1]{Address for Correspondence: Department of Statistics, University of Illinois at Urbana-Champaign, 605 E. Springfield Ave., Champaign, IL 61820 (Email: shuleiw@illinois.edu).}

\begin{abstract}
Recent advances in self-supervised learning have highlighted the efficacy of data augmentation in learning data representation from unlabeled data. Training a linear model atop these enhanced representations can yield an adept classifier. Despite the remarkable empirical performance, the underlying mechanisms that enable data augmentation to unravel nonlinear data structures into linearly separable representations remain elusive. This paper seeks to bridge this gap by investigating under what conditions learned representations can linearly separate manifolds when data is drawn from a multi-manifold model. Our investigation reveals that data augmentation offers additional information beyond observed data and can thus improve the information-theoretic optimal rate of linear separation capacity. In particular, we show that self-supervised learning can linearly separate manifolds with a smaller distance than unsupervised learning, underscoring the additional benefits of data augmentation. Our theoretical analysis further underscores that the performance of downstream linear classifiers primarily hinges on the linear separability of data representations rather than the size of the labeled data set, reaffirming the viability of constructing efficient classifiers with limited labeled data amid an expansive unlabeled data set.
\end{abstract}



\newpage
\section{Introduction}
\label{sc:intro}

\subsection{Self-Supervised Representation Learning}
The recent advance in large-scale machine learning models demonstrates their superior capacity and performance in various fields \citep{vaswani2017attention}, but also demands millions of labeled samples for training, which can be inaccessible in some applications \cite{dosovitskiyimage}. Self-supervised pre-training and transfer learning are introduced to address the challenge of scarce labeled data in natural language processing and computer vision. Unlike the classical supervised learning framework, the current self-supervised learning framework usually involves two steps: self-supervised pre-training and fine-tuning in the downstream task. In the pre-training stage, self-supervised learning first learns data representations/features from a large unlabeled dataset and pseudo-labels automatically generated from the unlabeled data \citep{hjelm2018learning,bachman2019learning,chen2020simple,he2020momentum,chen2021exploring,zbontar2021barlow,he2022masked}. The pre-trained representations are then transferred to train a full model via fine-tuning on a labeled dataset in the downstream task \citep{kumar2022fine}. A comprehensive review is available in \cite{balestriero2023cookbook}. Recent progress shows that self-supervised pre-training can reduce the need for external supervision and extract information from a large amount of unlabeled data more efficiently than classical unsupervised learning methods \citep{lecun2022path}. Due to these beneficial properties, self-supervised learning has been widely used in the pre-training of large foundational models \citep{zhou2023comprehensive}. For instance, self-supervised pre-training has played a crucial role in the success of recent large language models, such as BERT \citep{kenton2019bert} and GPT \citep{radford2018improving,radford2019language}.

Generating pseudo labels from unlabeled data is a pivotal component of self-supervised representation learning and a significant departure from traditional supervised and unsupervised learning methods. Various approaches to pseudo label generation are employed in different applications of self-supervised pre-training, including data masking \citep{brown2020language,he2022masked}, data augmentation \citep{chen2020simple,zbontar2021barlow}, and multi-modal data fusion \citep{akbari2021vatt}. These generated pseudo labels serve as supervisory signals in self-supervised representation learning, with the trained representations subsequently fine-tuned for diverse downstream tasks. Notably, empirical observations suggest that linear probing of learned representations, achieved by training a linear model on top of frozen representations, can surpass training all parameters of the same model from scratch \citep{chen2020simple,chen2020improved,kumar2022fine}. This implies that the representations acquired through self-supervised pre-training exhibit linear separability, enhancing downstream task efficiency via a linear model. Given these intriguing empirical findings, questions arise about why training with pseudo labels yields linearly separable representations and what underlying structural information self-supervised representation learning captures through pseudo labels. This paper endeavors to address these questions by theoretically investigating the linear separation capacity of self-supervised representation learning when pseudo labels are generated using data augmentation.

Motivated by recent empirical successes, numerous existing studies have embarked on exploring the theoretical foundations of self-supervised pre-training's advantages in downstream tasks \citep{tsai2020self,tian2020understanding,wen2022mechanism,saunshi2022understanding,cabannes2023ssl}. Specifically, a prevalent approach to investigating self-supervised learning is grounded in the context of linear representation or linear prediction functions \citep{tosh2021contrastive,lee2021predicting,wen2021toward,wang2022self,kumar2022fine}. While the linear framework facilitates precise characterization and insightful insights in this specific scenario, it may fall short of comprehensively elucidating the triumph of self-supervised learning, given the inherently nonlinear nature of data distributions. In addition to linear structures, existing endeavors have embraced more adaptable models, including conditional distributions of discrete latent variables \citep{arora2019theoretical}, distributions on graphs \citep{wei2020theoretical,balestriero2022contrastive}, nonlinear representations via kernels \citep{johnson2022contrastive}, and manifolds \citep{haochen2021provable,wang2022augmentation}. Despite the remarkable progress, the mechanisms behind how and why pseudo labels facilitate the transformation of nonlinear structures into linearly separable representations remain enigmatic. To bridge this gap, our study delves into self-supervised learning within a multi-manifold model, demonstrating the pivotal role of pseudo labels in learning linearly separable representations, particularly when manifolds are close to each other.

\subsection{Linearly Separable Representation and Unsupervised Learning}
In this and the next subsections, we demonstrate the key idea and results using a two-manifold case, with a more generalized case and formal results provided in later sections. Specifically, we consider the scenario where the observed data is drawn from a union of two smooth and compact $d$-dimensional manifolds:
$$
\Mcal = \Mcal_+ \cup \Mcal_-,
$$
where $\Mcal_+$ and $\Mcal_-$ correspond to distinct classes. The objective is to develop a classifier that distinguishes data originating from these two manifolds. In representation learning, our aim is to construct a mapping $\Theta: \Mcal \to \RR^S$ from the observed data, enabling us to build a more effective classifier using $\Theta(x)$ as opposed to the conventional data representation $x$. Recent advancements in self-supervised representation learning have indicated that a simple linear classifier can achieve perfect accuracy in downstream tasks when a linearly separable representation $\Theta(x)$ can be learned. Formally, the data representation $\Theta(x)$ is considered linearly separable if there exists a weight vector $w \in \RR^S$ satisfying
$$
\sup_{x \in \Mcal_+} w^T \Theta(x) < \inf_{x \in \Mcal_-} w^T \Theta(x).
$$
Given a dataset $(X_1, \ldots, X_n)$ randomly drawn from $\Mcal$, it is natural to inquire about the conditions under which consistent learning of linearly separable representations is possible, as well as the potential benefits of using pseudo labels to enhance the learning process.

To address these questions, we first analyze the performance of the graph Laplacian-based method, a classical unsupervised learning approach widely used in spectral clustering methods \citep{ng2001spectral, belkin2003laplacian, coifman2006diffusion,von2007tutorial,arias2017spectral,trillos2021large,chen2021diffusion}. Our investigation reveals that the resulting representation converges to a linearly separable one when the manifolds are well-separated, as indicated by the condition
\begin{equation}
	\label{eq:upml}
	\delta(\Mcal)\gg \left(\log n \over n\right)^{1/d},
\end{equation}
where $\delta(\Mcal)$ represents the smallest Euclidean distance between $\Mcal_+$ and $\Mcal_-$, given by
$$
\delta(\Mcal):=\inf_{x\in \Mcal_{+}, y\in \Mcal_{-}}\|x-y\|.
$$
When the condition \eqref{eq:upml} is not met, our analysis demonstrates that there exists an instance in which the graph Laplacian-based method treats $\Mcal$ as a single manifold rather than a union of two separated manifolds. Furthermore, we establish an information-theoretic lower bound, revealing that any method fundamentally struggles to differentiate a single manifold from a union of multiple manifolds with
$$
\delta(\Mcal)\le c\left(\log n \over n\right)^{1/d},
$$
where $c$ is a sufficiently small constant. These findings indicate that condition \eqref{eq:upml} serves as a necessary and sufficient requirement for learning linearly separable representation in the context of observing an unlabeled data set. Importantly, the graph Laplacian-based method achieves rate-optimal linear separation capacity in the two-manifold setting. This leads us to the question of whether there are strategies to enhance the linear separation capacity stated in \eqref{eq:upml} by leveraging additional information.

\subsection{Data Augmentation Improves Linear Separation Capacity}
\begin{figure}[h!]
	\begin{center}
		\begin{tikzpicture}
			\draw[thick,black](0,0)node{\includegraphics[width=0.4\textwidth]{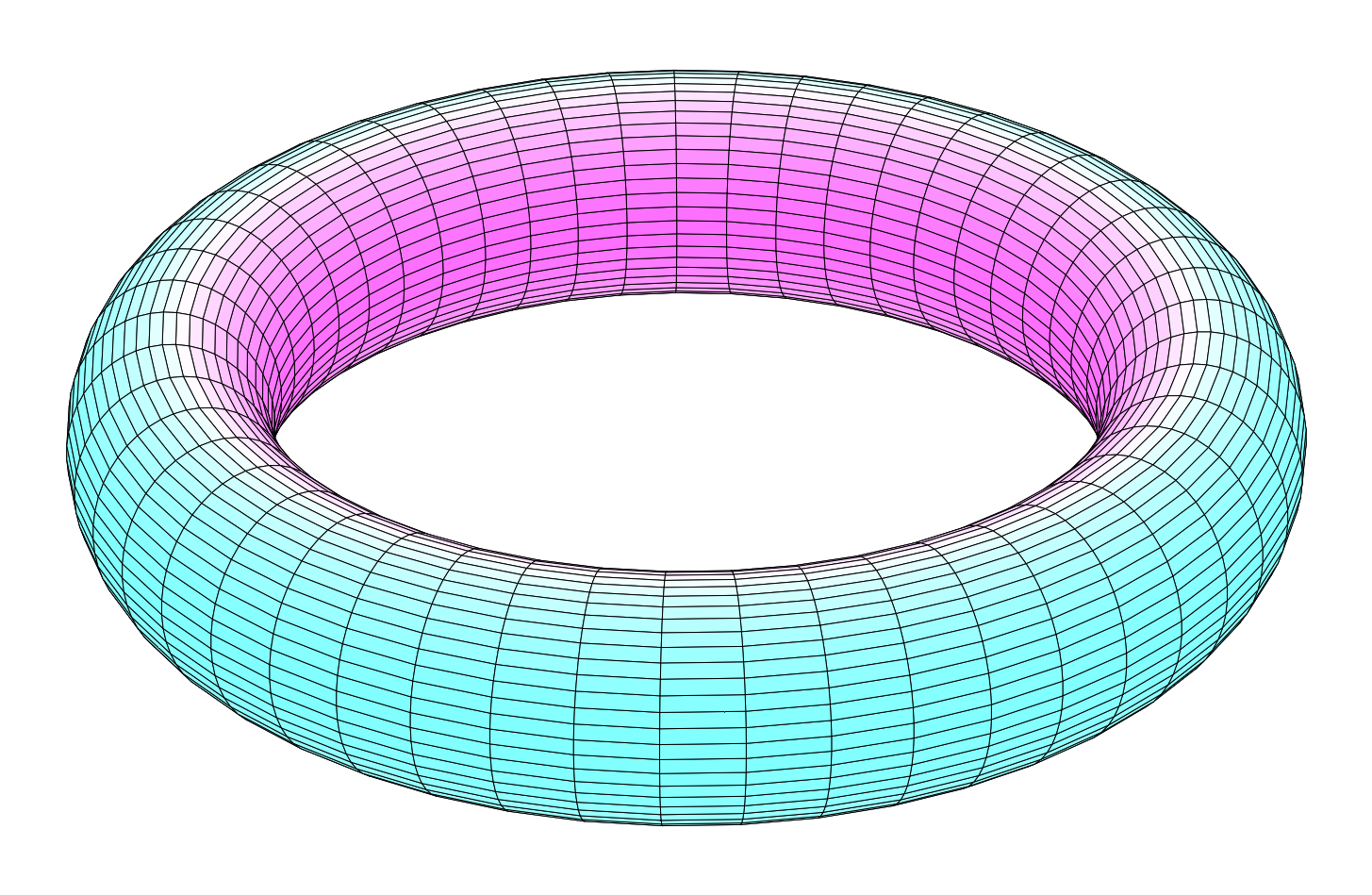}};
			\path[-latex,draw,line width=1pt] (1.75,0.4) -- (4.5,-0.8);
			\path[-latex,draw,line width=1pt] (2,1.05) -- (4.5,0.8);
			\draw[thick,black](5.2,-0.8)node{\includegraphics[width=0.07\textwidth]{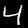}};
			\draw[thick,black](5.2,0.8)node{\includegraphics[width=0.07\textwidth]{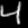}};
			\path[-latex,draw,line width=1pt] (-2,-1.05) -- (-4.5,-0.8);
			\path[-latex,draw,line width=1pt] (-1.75,-0.4) -- (-4.5,0.8);
			\draw[thick,black](-5.2,-0.8)node{\includegraphics[width=0.07\textwidth]{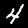}};
			\draw[thick,black](-5.2,0.8)node{\includegraphics[width=0.07\textwidth]{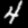}};
			\draw[-latex,thick,text=black](-5.9,-0.8) to[out=135,in=-135] (-5.9,0.8);
			\draw[thick,black](-6.6,0)node{\rotatebox{90}{\scriptsize Data Augmentation}};
			\draw[-latex,thick,text=black](5.9,0.8) to[out=-45,in=45] (5.9,-0.8);
			\draw[thick,black](6.6,0)node{\rotatebox{270}{\scriptsize Data Augmentation}};
		\end{tikzpicture}
	\end{center}
\caption{A toy example of a single product manifold: the circle with major radius captures data augmentation invariant structure, and the circle with minor radius captures irrelevant structure due to data augmentation. Data augmentation can help randomly draw samples from each smaller circle.}
\label{fg:productmanifold}
\end{figure}
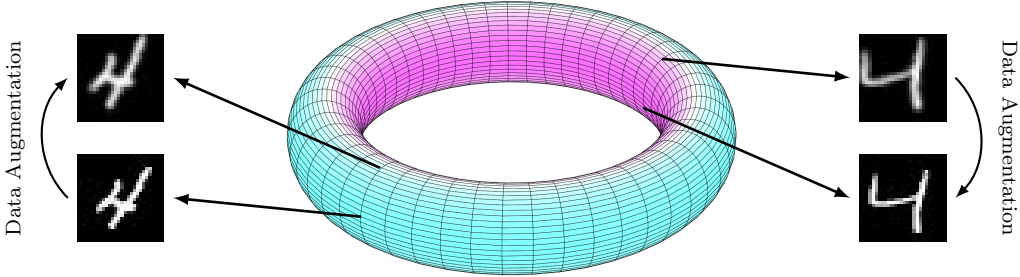

The graph Laplacian-based method is efficient at separating two manifolds but often overlooks a rich source of invariant structures present in the observed data across many applications. For instance, an image can represent the same context and be considered equivalent even after applying data augmentation techniques such as flipping, rotation, and cropping \citep{shorten2019survey, chen2020group}. To capture such invariant structures in the data, we can extend our assumption that each $\Mcal_+$ and $\Mcal_-$ is an isometric embedding of a product manifold:
$$
\Mcal_+=T_+(\Ncal_{s,+}\times \Ncal_{v,+}) \qquad \text{and} \qquad \Mcal_-=T_-(\Ncal_{s,-}\times \Ncal_{v,-}),
$$
where $T_+$ and $T_-$ are isometric diffeomorphisms. Here, $\Ncal_{s,+}$ and $\Ncal_{s,-}$ represent the $d_s$-dimensional manifold capturing the data augmentation invariant structure, while $\Ncal_{v,+}$ and $\Ncal_{v,-}$ correspond to the $d_v$-dimensional manifolds related to irrelevant structure due to data augmentation. A similar product manifold model is also employed in \cite{lederman2018learning,wang2022augmentation}. A toy example of a single product manifold is illustrated in Figure~\ref{fg:productmanifold}. This model offers a straightforward approach to modeling the data augmentation process. For instance, when $X_i\in \Mcal_+$, we can express the data point as $X_i=T_+(\phi_i,\psi_i)$, where $\phi_i\in \Ncal_{s,+}$ and $\psi_i\in \Ncal_{v,+}$, and describe its augmented data as $X'_i=T_+(\phi_i,\psi'_i)$, where $\psi'_i$ is randomly drawn from $\Ncal_{v,+}$. Similarly, when $X_i\in \Mcal_-$, $\psi'_i$ is drawn randomly from $\Ncal_{v,-}$.

Unlike classical unsupervised techniques, self-supervised learning explores the aforementioned invariant structure of observed data when pseudo labels are generated by data augmentation. Specifically, the representations learned by most existing data-augmentation based self-supervised learning methods aim to preserve similarity between augmented data, i.e., $\Theta(X_i)\approx \Theta(X'_i)$. Can we better learn a linearly separable representation by exploiting the similarity of augmented data? To address this question, this paper studies the performance of Augmentation Invariant Manifold Learning (AIML), introduced in \cite{wang2022augmentation}, as it simultaneously leverages the manifold's low dimensionality and the structure induced by augmented data. Capturing the invariant structure in data doesn't appear directly related to linear separation capacity,  bacause it's often believed that data augmentation helps to reduce redundant information \citep{chen2020group, wang2022self}. However, our investigation suggests that augmentation invariant manifold learning can surprisingly yield a linearly separable representation by requiring a smaller distance between manifolds than the graph Laplacian-based method, i.e., 
\begin{equation}
	\label{eq:upaiml}
	\delta(\Mcal)\gg \left(\log n \over n\right)^{1/d_s}.
\end{equation}
In other words, the linear separation capacity of augmentation invariant manifold learning relies solely on the dimension of the data augmentation invariant structure. These results suggest that selecting a data augmentation method that maintains a concise invariant structure (i.e., smaller $d_s$) can better facilitate the learning of a linearly separable representation. When data augmentation can be applied, our analysis further demonstrates that no method can consistently discern whether the observed data is drawn from a single manifold or a union of multiple manifolds with 
$$
\delta(\Mcal)\le c\left(\log n \over n\right)^{1/d_s}
$$
for some sufficiently small constant $c$. Combined with the upper bound in \eqref{eq:upaiml}, we observe that augmentation invariant manifold learning can achieve rate-optimal linear separation capacity when exploring the invariant structure within observed data through data augmentation.

\subsection{Impact on Downstream Classifier}
If we compare unsupervised and self-supervised learning, data augmentation can improve the optimal rate of linear separation capacity by
$$
\left(\log n \over n\right)^{1/d} \qquad \Rightarrow \qquad \left(\log n \over n\right)^{1/d_s}.
$$
In other words, through data augmentation, self-supervised learning can achieve separation between manifolds with a finer resolution, detecting differences between manifolds in greater detail compared to unsupervised learning. However, how does this improved linear separation capacity impact downstream analysis?

To investigate the impact on downstream analysis, we consider training a logistic regression, one of the most widely used linear classifiers, on top of learned representations to predict whether data originates from $\Mcal_+$ or $\Mcal_-$. The logistic regression is trained using gradient descent on a limited labeled dataset. Our investigation reveals that the misclassification rate of the resulting logistic classifier primarily hinges on how effectively the learned data representation linearly separates the data, rather than the size of the labeled dataset in downstream tasks. In other words, training on a modest number of labeled samples can lead to a precise linear classifier, as long as the nonlinear structure is disentangled by the data representation learned from unlabeled samples. Consequently, with the aid of data augmentation, the data representation learned through self-supervised learning can result in a superior linear classifier for downstream analysis compared to unsupervised learning. These theoretical findings are also validated through numerical examples.

\section{Model and Linear Separable Representation}
\label{sc:model}

In this section, we generalize the two-manifold setting in the Introduction section to a multi-manifold model.

\subsection{A Multi-Manifold Model for Data Augmentation}
The conventional representation of data is often in a high-dimensional space, but empirical observations suggest that the data distribution in various applications, such as natural image data \citep{pope2021intrinsic}, can be characterized by latent low-dimensional variables. In particular, the manifold assumption is commonly used to model the underlying low-dimensional structure in high-dimensional space. Alongside the low-dimensional structure, data can exhibit grouping into small clusters \citep{martinez2001statistics,basri2003lambertian,fu2005similarity,vidal2006unified}. Motivated by these insights, we consider data drawn from a union of multiple manifolds
$$
\Mcal=\Mcal_1\cup\Mcal_2\cup\ldots\cup\Mcal_K,
$$
where each $\Mcal_k\subset \RR^D$ is a smooth and compact manifold of dimension $d_k$ representing a distinct subclass of data. Each manifold $\Mcal_k$ may correspond to a subset of a category or class, such as images of blue shirts, blue dresses, red shirts, and red dresses. We assume that the manifolds $\Mcal_1,\ldots,\Mcal_K$ are well-separated, meaning that
$$
\delta(\Mcal):=\min_{1\le k_1<k_2\le K}\left(\inf_{x\in \Mcal_{k_1}, y\in \Mcal_{k_2}}\|x-y\|\right)>0.
$$
Here, $\delta(\Mcal)$ represents the smallest Euclidean distance between any pair of manifolds. Since data augmentation transformations are believed not to change the subclass of data \citep{chen2020group}, we adopt a similar approach as in \cite{wang2022augmentation} by assuming each $\Mcal_k$ is an isometric embedding of a product manifold (Figure~\ref{fg:productmanifold})
$$
\Mcal_k=T_k(\Ncal_{s,k}\times \Ncal_{v,k}),
$$
where $T_k$ is an isometric diffeomorphism, and $\Ncal_{s,k}, \Ncal_{v,k}$ are two manifolds with dimensions $d_{s,k}$ and $d_{v,k}$, such that $d_k=d_{s,k}+d_{v,k}$. Here, $\Ncal_{s,k}$ and $\Ncal_{v,k}$ represent $\Mcal_k$'s the invariant structure of interest and irrelevant nuisance structures resulting from data augmentation, respectively. For simplicity, we assume $d_{s,1}=\ldots=d_{s,K}=d_s$, $d_{v,1}=\ldots=d_{v,K}=d_v$, and $d_{1}=\ldots=d_{K}=d$.

Given the above multi-manifold assumption, we need to define how an observed data point $X$ is sampled from $\Mcal$ and how the augmented data $X'$ is sampled given a data point $X$. We assume that the observed data point $X$ is drawn from a mixture distribution $\pi$ defined on $\Mcal$:
$$
	\pi(x) = \sum_{k=1}^K w_k \pi_k(x), \quad \text{where} \quad w_k > 0 \quad \text{and} \quad \sum_{k=1}^K w_k = 1.
$$
Here, $\pi(x)$ and $\pi_k(x)$ are probability density functions defined on $\Mcal$, with $\pi_k(x)$ having support only on $\Mcal_k$. In other words, $X$ is drawn from $\Mcal_k$ with probability $w_k$. Next, we discuss how data are sampled from $\pi_k$ on each $\Mcal_k$. The product manifold assumption suggests that all points on $\Mcal_k$ can be represented as $X = T_k(\phi, \psi)$, where $\phi \in \Ncal_{s,k}$ and $\psi \in \Ncal_{v,k}$. Thus, it suffices to consider the probability distribution of $\phi$ and $\psi$ on $\Ncal_{s,k}$ and $\Ncal_{v,k}$. We sample independent random variables $\phi \in \Ncal_{s,k}$ and $\psi \in \Ncal_{v,k}$ as follows:
$$
	\phi \sim \pi_k^s(\phi) \quad \text{and} \quad \psi \sim \pi_k^v(\psi).
$$
Then, we set $X = T_k(\phi, \psi)$. For simplicity, we assume that $\pi_k^v$ is a uniform distribution on $\Ncal_{v,k}$. The product manifold assumption also provides a straightforward way to model the sampling process of augmented data $X'$ given a data point $X$. Given $X = T_k(\phi, \psi) \in \Mcal_k$, we assume that the augmented data $X' = T_k(\phi, \psi')$, where $\psi'$ is another independent random variable drawn from $\pi_k^v$ on $\Ncal_{v,k}$. In other words, given $X = T_k(\phi, \psi) \in \Mcal_k$, we randomly sample the augmented data point from the fiber $\Mcal(\phi) = \{x \in \Mcal_k: x = T_k(\phi, \psi), \psi \in \Ncal_{v,k}\}$.

\subsection{Linearly Separable Representation}
Representation learning aims to learn a mapping $\Theta: \Mcal\to \RR^S$ that enhances the performance of downstream analyses, including clustering, classification, and regression. In this paper, we focus solely on classification as the downstream task. In a classification task, our goal is to train a classifier $H:\Mcal\to \{-1,1\}$ that predicts the true label $H^\ast:\Mcal\to \{-1,1\}$, where $-1$ and $1$ denote the two possible binary label values. Among the simplest and most commonly used classifiers is the linear classifier:
$$
H_{w,\theta}(x)=\begin{cases}
	1 & w^Tx >\zeta\\
	-1& w^Tx \le \zeta
\end{cases},
$$
where $w\in \RR^D$ represents the weight vector and $\zeta\in \RR$ is a scalar threshold. Various algorithms have been proposed in the literature to train a linear classifier, such as logistic regression, linear discriminant analysis, and support vector machines \citep{hastie2009elements}. A linear classifier can achieve 100\% accuracy when the sets $\Mcal_+(H^\ast)=\{x\in\Mcal: H^\ast(x)=1\}$ and $\Mcal_-(H^\ast)=\{x\in\Mcal: H^\ast(x)=-1\}$ are linearly separable in terms of their $x$ values. However, in practice, $\Mcal_+(H^\ast)$ and $\Mcal_-(H^\ast)$ are often not linearly separable due to their complex shapes in high-dimensional space.

Through representation learning, our aim is to enhance the classifier trained by $\Theta(x)$, making it more accurate than the classifier trained using $x$. Additionally, we aspire for the data representation $\Theta(x)$ to be versatile, applicable across multiple classification labels. Specifically, we address a collection of classification labels within the multi-manifold model
$$
	\Hcal^\ast=\left\{H^\ast: H^\ast(x)\in \{-1,1\}, H^\ast(x)=H^\ast(y),\quad \forall \ x,y\in \Mcal_k,\quad 1\le k\le K \right\}.
$$
In this definition, we enforce that the labels of data points within each subclass, i.e., manifold $\Mcal_k$, are identical. As a result, $\Hcal^\ast$ encompasses $2^K$ distinct possible classification labels. Our objective is to develop a data representation that enhances the classifier's performance across all classification tasks within $\Hcal^\ast$. For the purpose of this study, we concentrate on a linear classifier and endeavor to construct a linearly separable representation. In particular, a data representation $\Theta(x)$ is deemed linearly separable if, for any classification task $H^\ast\in \Hcal^\ast$, a weight vector $w_{H^\ast}\in \RR^S$ can be found such that
$$
	\sup_{x\in\Mcal_+(H^\ast) }w_{H^\ast}^T\Theta(x) < \inf_{x\in\Mcal_-(H^\ast)}w_{H^\ast}^T\Theta(x).
$$
Having introduced this definition, the natural inquiry arises: does a linearly separable representation exist, and if so, how can we learn it from our observed data?

\section{Classical graph Laplacian-based Method}
\label{sc:manifoldlearning}
To learn data representation on the multi-manifold, it is sufficient to find a way to separate each individual manifold. A commonly used unsupervised method to learn individual manifold structure is based on the spectral analysis of the graph Laplacian on a neighborhood graph \citep{belkin2003laplacian,coifman2006diffusion}. Let $X_1,\ldots, X_n$ be independent identically distributed samples drawn from the distribution $\pi$, which is introduced in Section~\ref{sc:model}. Given the observed samples, we can construct a neighborhood graph by connecting an edge between two sample points $X_i$ and $X_j$ if and only if $\|X_i-X_j\|\le r$, and assigning weights $W_{i,j}=\bI(\|X_i-X_j\|\le r)$, where $\bI(\cdot)$ is the indicator function. With the neighborhood graph introduced, we define the graph Laplacian matrix $L$ as
$$
L_{i,j}=\begin{cases}
	\sum_{i'\ne i}W_{i,i'} & \text{if } i=j\\
	-W_{i,j} & \text{if } i\ne j
\end{cases}.
$$
Then, the first $S$ eigenvectors (corresponding to the $S$ smallest eigenvalues) of $L$ (denoted $U_1,\ldots,U_S$) can form the data representations for $X_1,\ldots, X_n$. Since these data representations can be used to separate manifolds, they have already been employed in various versions of spectral clustering algorithms \citep{ng2001spectral,von2007tutorial,arias2017spectral,trillos2021large,chen2021diffusion}.

To understand why and when the graph Laplacian can help recover the multi-manifold structure, we need to study the asymptotic behavior of $L$'s eigenvectors and introduce the continuum level of the Laplacian operator. Specifically, we define the Laplacian operator on $\Mcal_k$ as
$$
\Delta_{\Mcal_k}\theta_k=-{1\over \pi_k}{\rm div}_{\Mcal_k}(\pi_k^2\nabla_{\Mcal_k}\theta_k),
$$
where $\theta_k: \Mcal_k \to \RR$ is a function defined on $\Mcal_k$. After introducing Laplacian operator on each $\Mcal_k$, we now define the tensorized Laplacian operator $\Delta_{\Mcal}$
$$
\Delta_{\Mcal}\theta=(w_1\Delta_{\Mcal_1}\theta_1,\ldots, w_K\Delta_{\Mcal_K}\theta_K),
$$
where $\theta:\Mcal\to \RR$ is a function defined on $\Mcal$ and can be written as $\theta=(\theta_1,\ldots,\theta_K)$ where each $\theta_k$ is defined on $\Mcal_k$. Since the operator $\Delta_{\Mcal}$ is defined in a manifold-wise fashion, the eigenfunctions of $\Delta_{\Mcal}$ only has support on one manifold, i.e., the eigenfunctions $\theta_{k,l_k}(x)=\theta^k_{l_k}(x)/\sqrt{w_k}$ if $x\in \Mcal_k$ and $\theta_{k,l_k}(x)=0$ if $x\notin \Mcal_k$, where $\theta^k_{l_k}(x)$ is the $l_k$th eigenfunction of $\Delta_{\Mcal_k}$. In particular, the first $K$ eigenfunctions of $\Delta_{\Mcal}$ has the form $\theta(x)=c_k\bI(x\in\Mcal_k)$, where $c_k$ is some normalization constant and $1\le k\le K$. In other words, the structure information of different manifolds is mapped to different coordinates with the help of these eigenfunctions. Because of this special property, the representation based on these eigenfunctions can localize each manifold and thus lead to linearly separable representation. It is also interesting to note that these eigenfunctions can capture the geometric information within each manifold.

The primary reason behind the capability of the eigenvectors of $L$ to recover the multi-manifold structure stems from the fact that the graph Laplacian provides a reliable approximation of the Laplacian operator $\Delta_{\Mcal}$, and the eigenvectors of $L$ converge to the eigenfunctions of $\Delta_{\Mcal}$. To establish this convergence, we need to introduce the following assumption:
\begin{assumption}
	\label{ap:ml}
	We assume that the following conditions hold:
	\begin{enumerate}
		\item There exists a constant $C_\pi>1$ such that
		$$
		{1\over C_\pi}\le \pi_k^s(\phi)\le C_\pi, \quad 1\le k\le K;
		$$
		\item The parameter $r$ is chosen such that 
		$$
		2r<\min \{1,i_o,\Gamma^{-1/2},R/2\},
		$$
		where $R$ and $\Gamma$ denote the upper bounds of the reach and absolute values of sectional curvatures, respectively, and $i_0$ is a lower bound on the injectivity radius. 
	\end{enumerate}
\end{assumption}
These assumptions are commonly used in the analysis of the convergence of the spectrum of the graph Laplacian \citep{calder2022improved,trillos2021large}. Given Assumption~\ref{ap:ml}, we can establish the convergence of the eigenvectors of $L$.
\begin{theorem}
	\label{thm:upperml}
	If Assumption~\ref{ap:ml} holds and we assume $r\to 0$,
	\begin{equation}
		\label{eq:mlcondition}
	r\gg \left(\log n\over n\right)^{1/d}\qquad {\rm and}\qquad \delta(\Mcal)>r,
\end{equation}
	then with probability at least $1-4Kn^{-\alpha}$, if $U_s$ is normalized eigenvector of $L$ with eigenvalue $\lambda_s(L)$, there is a normalized eigenfunction $\theta_s$ of $\Delta_{\Mcal}$ with eigenvalue $\lambda_s(\Mcal)$ such that
	$$
	\|U_s-\vec{\theta}_s\|_{L^2(\pi_n)}:=\sqrt{{1\over n}\sum_{i=1}^n\left((U_s(X_i)-\theta_s(X_i)\right)^2}\to 0,
	$$
	where $\vec{\theta}_s=(\theta_s(X_1),\ldots, \theta_s(X_n))\in \RR^n$.
\end{theorem}
The proof of Theorem~\ref{thm:upperml} adopts the same variational method as in \cite{burago2015graph,garcia2020error,calder2022improved,trillos2021large}. The detailed convergence rate of eigenfunctions can be found in Appendix. Theorem~\ref{thm:upperml} shows that under certain conditions, including the main condition \eqref{eq:mlcondition}, the eigenvectors of $L$ can approximate the eigenfunctions of $\Delta_{\Mcal}$ very well and thus lead to linearly separable representation. Can these conditions in \eqref{eq:mlcondition} be relaxed? The lower bound condition for $r$ in \eqref{eq:mlcondition} is sharp since the connectivity threshold of the random geometric graph is at order $(\log n/n)^{1/d}$ \citep{penrose2003random}. The other condition $\delta(\Mcal)>r$ is also a necessary condition for separating different manifolds. To show this, we present the following example and theorem to argue that the eigenvectors of $L$ cannot separate multi-manifold when $\delta(\Mcal)\ll r$.

Consider a simple example of two manifolds $\Mcal=\Mcal_1\cup \Mcal_2$. $\Mcal_1$ and $\Mcal_2$ are defined in the following way:
$$
\Mcal_1=\left\{(y,0): y\in \tilde{\Mcal}\right\}\qquad {\rm and}\qquad \Mcal_2=\left\{(y,z^o): y\in \tilde{\Mcal}\right\},
$$
where $\tilde{\Mcal}$ is a $d$-dimensional smooth and compact manifold embedded in $\RR^{D-1}$, and $z^o\in \RR$ is a positive constant. This construction clearly shows that $\delta(\Mcal)=z^o$. The following theorem shows that eigenvectors of $L$ converge to eigenfunctions of $\Delta_{\tilde{\Mcal}}$ instead of $\Delta_{\Mcal}$.
\begin{theorem}
	\label{thm:lowerml}
	Suppose $w_1=w_2=1/2$, and $\pi_1(x)$ and $\pi_2(x)$ are the uniform distribution on $\Mcal_1$ and $\Mcal_2$ in above example. So we can write our observed data as $X_i=(Y_i,Z_i)$ for $1\le i\le n$, where $Y_i\in \RR^{D-1}$ is drawn from a uniform distribution on $\tilde{\Mcal}$ and $\PP(Z_i=z^o)=\PP(Z_i=0)=1/2$. If Assumption~\ref{ap:ml} holds and we assume $r\to 0$,
	$$
	r\gg \left(\log n\over n\right)^{1/d}\qquad {\rm and}\qquad \delta(\Mcal)\ll r,
	$$
	then with probability at least $1-Cn^{-2}$ for some constant $C$, if $U_s$ is normalized eigenvector of $L$ with $s$th eigenvalue, there is a normalized eigenfunction $\theta_s$ of $\Delta_{\tilde{\Mcal}}$ with $s$th eigenvalue such that
	$$
	\|U_s-\vec{\theta}_s\|_{L^2(\pi_n)}\to 0,
	$$
	where $\vec{\theta}_s=(\theta_s(Y_1),\ldots, \theta_s(Y_n))\in \RR^n$.
\end{theorem}
As $\vec{\theta}_s$ depends solely on $Y_1,\ldots, Y_n$ and not on $Z_1,\ldots,Z_n$, Theorem~\ref{thm:lowerml} suggests that the eigenvectors of $L$ cannot effectively distinguish between $\Mcal_1$ and $\Mcal_2$, thereby failing to provide a linearly separable representation when $\delta(\Mcal)\ll r$. The combined implications of Theorem~\ref{thm:upperml} and \ref{thm:lowerml} imply that the classical Laplacian-based method necessitates well-separated manifolds in order to learn a linearly separable representation, meaning that the minimum distance between manifolds must be sufficiently large:
$$
\delta(\Mcal)\gg \left(\log n \over n\right)^{1/d}.
$$
While this minimum distance requirement optimally applies to the classical graph Laplacian-based method, it naturally raises the question of whether alternative methods can effectively separate closely situated manifolds.

We investigate the information-theoretic lower bound for separating manifolds to answer this question. Following a similar strategy in \cite{mossel2015reconstruction}, we address a simpler question: what is the smallest distance between manifolds that enables us to distinguish whether our observed data is drawn from one or two manifolds? Specifically, we consider the following hypothesis testing problem:
\begin{equation}
	\label{eq:hypothesis}
	\HH_0: \Mcal=\Mcal_0\qquad {\rm v.s.}\qquad \HH_1: \Mcal=\Mcal_1\cup\Mcal_2.
\end{equation}
Here $\Mcal_0$, $\Mcal_1$, and $\Mcal_2$ are smooth and compact $d$-dimensional manifolds such that the distance between $\Mcal_1$ and $\Mcal_2$ is $\delta(\Mcal)>0$. In other words, under the null hypothesis, the data is drawn from a single manifold, while under the alternative hypothesis, it is drawn from a union of two manifolds. Testing hypothesis in \eqref{eq:hypothesis} is a simpler problem than constructing linear separable representation because we can distinguish $\HH_0$ from $\HH_1$ if we can linearly separate $\Mcal_1$ and $\Mcal_2$. To detect the above hypothesis, we can consider a test $T:(X_1,\ldots, X_n)\to \{0,1\}$ such that we reject the null hypothesis if $T=1$. A test $T$ is called $\alpha$-level test if $\PP_0(T=1)\le \alpha$ where $\PP_0$ is the probability measure under $\HH_0$. The following theorem characterizes the lower bound for separating manifolds.
\begin{theorem}
	\label{thm:lowergeneral}
	For any $\alpha$-level test $T(X_1,\ldots, X_n)$, there exists an instance where 
	$$
	\delta(\Mcal)\ge \left(b\log n \over n\right)^{1/d},
	$$
	for some small enough constant $b>0$, such that the type II error of this test converges to $1-\alpha$.
\end{theorem}
This theorem demonstrates that any method is fundamentally limited in its ability to separate manifolds when the minimum distance requirement is not met. Combining this result with Theorem~\ref{thm:upperml}, we observe that the optimal rate of linear separation capacity is given by
$$
\delta(\Mcal) \asymp \left(\log n \over n\right)^{1/d}.
$$
While the classical graph Laplacian-based method can achieve this optimal rate, a natural question arises: Can we uncover additional data structure and relax this minimum distance requirement by extracting more information from the data? 

\section{Augmentation Invariant Manifold Learning}
\label{sc:aiml}
Besides the observed samples, we can also generate additional augmented data from the observed data in various applications. For instance, cropping, rotation, colorization, and scaling can produce new images from the original images \citep{shorten2019survey}. The augmented data is treated as an equivalent yet distinct version of the original data, offering supplementary information beyond the observed data. In recent years, self-supervised representation learning methods, encompassing both contrastive and non-contrastive approaches, have been introduced to learn data representations by harnessing this equivalent relationship among augmented data \citep{chen2020simple,grill2020bootstrap,tian2020contrastive,chen2021exploring,zbontar2021barlow,wang2022augmentation}. The data augmentation invariant representations learned through these methods can enhance downstream analysis.

To leverage augmented data, existing self-supervised representation learning methods aim to learn a data representation that is invariant to augmented data, i.e.,
$$
	\Theta(X_i)\approx \Theta(X'_i),
$$
where $X'_i$ is a data point randomly generated by applying data augmentation techniques to $X_i$. Specifically, augmentation invariant manifold learning has been recently introduced to learn data representation by capturing the geometric structure of the manifold and the invariance property of augmented data \citep{wang2022augmentation}. To elaborate further, augmentation invariant manifold learning can be formulated as an stochastic optimization algorithm and the loss function on a small batch of samples $\Scal\subset\{1,\ldots,n\}$ is defined as
\begin{equation}
	\label{eq:aimlop}
	\hat{\ell}(\beta)=\sum_{i,j\in \Scal} W_{i,j}\|\Theta_\beta(X'_{i})-\Theta_\beta(X''_{j})\|^2+\lambda_1 \sum_{i\in \Scal}\|\Theta_\beta(X'_{i})-\Theta_\beta(X''_{i})\|^2+\lambda_2\Rcal(\Theta_\beta),
\end{equation}
where $X'_{i}$ and $X''_i$ are independent augmented copies of $X_i$, $W_{i,j}$ represents the weights between $X'_i$ and $X''_{j}$, and $\Rcal(\Theta_\beta)$ is a regularization term enforcing an orthonormal representation. In the above loss function, the first term corresponds to a computationally efficient version of the graph Laplacian, while the second term aims to capture the similarity between augmented data. The detailed algorithm of augmentation invariant manifold learning can be found in Algorithm~\ref{ag:aimlop} in Appendix.

\cite{wang2022augmentation} demonstrated that preserving the similarity of augmented data is equivalent to incorporating weights between augmented data of two samples. From this perspective, an equivalent form of augmentation invariant manifold learning in Algorithm~\ref{ag:aimlop} follows a similar procedure to the classical Laplacian-based method, but assesses the similarity between two samples using the average of weights between their augmented data:
$$
\bar{W}_{i,j}=\EE\left(\bI(\|X'_i-X'_j\|\le r)\right),
$$
where $X'_i$ and $X'_j$ are the independent augmented data of $X_i$ and $X_j$, and the expectation is taken over the data augmentation process. These new weights $\bar{W}_{i,j}$ lead to the construction of the graph Laplacian matrix $\bar{L}$ in the same manner as the classical Laplacian-based method. Consequently, the first $S$ eigenvectors of $\bar{L}$ serve as the new data representations. In scenarios involving a single manifold, \citep{wang2022augmentation} shows that augmentation invariant manifold learning can more effectively reduce dimensions nonlinearly compared to the classical Laplacian-based method. Given the resemblance between augmentation invariant manifold learning and the classical Laplacian-based method, a natural question emerges: Can augmentation invariant manifold learning also facilitate the learning of linearly separable representations? If so, how can data augmentation contribute to representation learning?

To study the properties of augmentation invariant manifold learning, it is necessary to introduce the Laplacian operator on the augmentation invariant manifolds $\Ncal_s := \Ncal_{s,1} \cup \ldots \cup \Ncal_{s,K}$. Given the multi-manifold $\Ncal_s$, we define a tensorized Laplacian operator $\Delta_{\Ncal_s}$ as follows:
$$
	\Delta_{\Ncal_s}\bar{\theta} = \left( \frac{w_1}{{\rm Vol}\Ncal_{v,1}} \Delta_{\Ncal_{s,1}}\bar{\theta}_1, \ldots, \frac{w_K}{{\rm Vol}\Ncal_{v,K}} \Delta_{\Ncal_{s,K}}\bar{\theta}_K \right),
$$
where $\bar{\theta}:\Ncal_s\to \RR$ is a function defined on $\Ncal_s$, $\bar{\theta}_k: \Ncal_{s,k} \to \RR$ is a function defined on $\Ncal_{s,k}$, ${\rm Vol}\Ncal_{v,k}$ represents the volume of $\Ncal_{v,k}$, and $\Delta_{\Ncal_{s,k}}$ denotes the Laplacian operator on $\Ncal_{s,k}$ given by
$$
	\Delta_{\Ncal_{s,k}}\bar{\theta}_k = -\frac{1}{\pi^s_k}{\rm div}_{\Ncal_{s,k}}({\pi^s_k}^2\nabla_{\Ncal_{s,k}}\bar{\theta}_k).
$$
Similar to $\Delta_{\Mcal}$, the eigenfunctions of $\Delta_{\Ncal_s}$ also have support on each individual manifold. Thus, a representation based on the eigenfunctions of $\Delta_{\Ncal_s}$ can effectively separate manifolds and lead to a linearly separable representation. In the following, we demonstrate that the eigenvectors of $\bar{L}$ converge to eigenfunctions of $\Delta_{\Ncal_s}$, enabling the detection of the multi-manifold structure.
\begin{theorem}
	\label{thm:upperaiml}
	If Assumption~\ref{ap:ml} holds and we assume $r\to 0$,
	\begin{equation}
		\label{eq:aimlcondition}
		r\gg \left(\log n\over n\right)^{1/d_s}\qquad {\rm and}\qquad \delta(\Mcal)>r,
	\end{equation}
	then with probability at least $1-4Kn^{-\alpha}$, if $\bar{U}_s$ is normalized eigenvector of $\bar{L}$ with $s$th eigenvalue, there is a normalized eigenfunction $\bar{\theta}_s$ of $\Delta_{\Ncal_s}$ with $s$th eigenvalue such that
	$$
	\|\bar{U}_s-\vec{\theta}_s\|_{L^2(\pi_n)}\to 0,
	$$
	where $\vec{\theta}_s=(\bar{\theta}_s(\phi_1),\ldots, \bar{\theta}_s(\phi_n))\in \RR^n$.
\end{theorem}
Theorem~\ref{thm:upperaiml} suggests that the graph Laplacian of $\bar{L}$ is a good approximation of the Laplacian operator $\Delta_{\Ncal_s}$ rather than $\Delta_{\Mcal}$. Thus, the eigenvectors of $\bar{L}$ can also detect each individual manifold and be used to construct a linearly separable representation when 
$$
\delta(\Mcal) \gg \left(\log n \over n\right)^{1/d_s}.
$$
In other words, the weight $\bar{W}_{i,j}$ defined by augmented data can separate manifolds with a smaller $\delta(\Mcal)$ than the classical Laplacian-based method. The intuition behind this improvement is that $\bar{W}_{i,j}$ measures the similarity between the fibers $\Mcal(\phi_i)$ and $\Mcal(\phi_j)$ rather than $X_i$ and $X_j$, leading a kernel between latent variables $\phi_i$ and $\phi_j$ even though we cannot observe $\phi_i$ and $\phi_j$ directly. More specifically, $\bar{W}_{i,j}$ can be approximately written as the following kernel 
$$
{1\over r^{d_v}}\bar{W}_{i,j} \approx {V_{d_v}\over {\rm Vol}\Ncal_{v,k}}\left(1-{d_{\Ncal_s}^2(\phi_i,\phi_j)\over r^{2}}\right)_+^{d_v/2},
$$
where $X_i, X_j \in \Mcal_k$, $V_{d_v}$ is the volume of a $d_v$-dimensional unit ball, and $(x)_+ = \max(x,0)$. Therefore, capturing the invariant data structure can help reduce dimension nonlinearly and separate different manifolds.

We show that the multiple manifolds can be better separated via capturing the invariant structure of data, but is the above minimum distance requirement sharp? To address this, we establish an information-theoretic lower bound for the task of separating manifolds by analyzing the hypothesis testing problem introduced in \eqref{eq:hypothesis}. Due to the presence of data augmentation, we are not limited to observing individual samples $X_1, \ldots, X_n$; instead, we have access to a collection of fiber sets $\Mcal(\phi_1), \ldots, \Mcal(\phi_n)$. When data augmentation is available, we can consider a test defined by these fibers, that is, $T:(\Mcal(\phi_1), \ldots, \Mcal(\phi_n))\to \{0,1\}$ such that we reject the null hypothesis if $T=1$. The ensuing theorem establishes a lower bound for the task of separating manifolds by leveraging the invariant structure of augmented data.
\begin{theorem}
	\label{thm:loweraiml}
	For any $\alpha$-level test $T(\Mcal(\phi_1),\ldots, \Mcal(\phi_n))$, there exists an instance where 
	$$
	\delta(\Mcal)\ge \left(b\log n \over n\right)^{1/d_s},
	$$
	for some small enough constant $b>0$, such that the type II error of this test converges to $1-\alpha$.
\end{theorem}

The results presented in Theorems~\ref{thm:upperaiml} and \ref{thm:loweraiml} demonstrate that augmentation invariant manifold learning can attain the optimal rate of linear separation capacity. By leveraging the additional information gained through data augmentation, augmentation invariant manifold learning enhances the rate of linear separation capacity to the following level:
$$
\delta(\Mcal) \asymp \left(\log n \over n\right)^{1/d_s}.
$$
A comparison between Theorem~\ref{thm:lowergeneral} and \ref{thm:loweraiml} underscores the indispensability of augmented data in acquiring linearly separable data representations, particularly when the manifolds exhibit proximity to one another.

\section{Impact on Downstream Linear Classifier}
\label{sc:downstream}
The previous sections characterize the required conditions for learning linearly separable representations, but it is still unclear how these representations can lead to an efficient linear classifier. To demystify the effectiveness of linearly separable representations, we consider a binary classification downstream task and study the performance of logistic regression. To be specific, suppose we observe a collection of labeled samples $(\tilde{X}_1,Y_1),\ldots, (\tilde{X}_m,Y_m)$ where the label $Y_i=H^\ast(\tilde{X}_i)$ for some $H^\ast\in \Hcal^\ast$. In the logistic regression, our goal is to minimize the following empirical logistic risk
$$
	\min_{\beta\in \RR^S} \quad \Lcal_{\rm log}(\beta)={1\over m}\sum_{i=1}^m\ln\left(1+\exp\left[-Y_i\beta^T\hat{\Theta}(\tilde{X}_i)\right]\right),
$$
where $\hat{\Theta}:\Mcal\to \RR^S$ is some data representations resulted from previous sections. A commonly used way to minimize $\Lcal_{\rm log}(\beta)$ is the gradient descent method, i.e.,
$$
\beta_{t+1}=\beta_t-\eta_t\nabla \Lcal_{\rm log}(\beta_t),
$$
where $\eta_t$ is the step size in the $t$th iteration. After stopping the iterations, the resulting classifier is 
$$
\hat{H}_{\hat{\Theta}, \hat{\beta}}(x)=\begin{cases}
	1,& \hat{\beta}^T\hat{\Theta}(x)>0\\
	-1,& \hat{\beta}^T\hat{\Theta}(x)\le0
\end{cases},
$$
where $\hat{\beta}$ is the resulting weighted vector in the above gradient descent iterates. To study the performance of $\hat{H}_{\hat{\Theta}, \hat{\beta}}$, we consider the misclassification rate as our measure
$$
\xi(\hat{\Theta})=\PP\left(\hat{H}_{\hat{\Theta}, \hat{\beta}}(X)\ne H^\ast(X)\right).
$$

To characterize the theoretical properties of $\hat{H}_{\hat{\Theta}, \hat{\beta}}$, we consider the following assumptions:
\begin{assumption}
	\label{ap:downstream}
	It holds that
	\begin{enumerate}
		\item $\tilde{X}_1,\ldots, \tilde{X}_m$ is independent of the data in unsupervised/self-supervised learning ($X_1, \ldots, X_n$), so $\tilde{X}_1,\ldots, \tilde{X}_m$ is independent of $\hat{\Theta}$. 
		\item We choose the data representation as the first $K$ eigenfunctions resulting from the classical Laplacian-based method or augmentation invariant manifold learning, i.e.,  $\hat{\Theta}(x)=(\hat{\theta}_1(x),\ldots, \hat{\theta}_K(x))$. We also assume $\hat{\theta}_s(x)$ is close to $\theta_s(x)$ in the following sense
		$$
		\sum_{k=1}^Kw_k\int_{\Mcal_k} (\hat{\theta}_s(x)-\theta_s(x))^2\pi_k(x)dx\le \chi_n,\qquad 1\le s\le K,
		$$
		where $\theta_s(x)$ is the $s$th eigenfunction of $\Delta_{\Mcal}$ or $\Delta_{\Ncal_s}$.
		\item For each $1\le s\le K$, we have at least one sample $\tilde{X}_i\in \Mcal_s$. 
		\item The step size of the gradient descent is chosen as $\eta_t=1$.
	\end{enumerate}
\end{assumption}
The above assumptions are mild in practice. In particular, the second assumption is just a population version of the results in Theorem~\ref{thm:upperml}, \ref{thm:lowerml} and \ref{thm:upperaiml}, and the last assumption is used for the convergence of the gradient descent method when the data is linearly separable \cite{soudry2018implicit,ji2018risk}. With these assumptions, the following theorem shows the convergence of the misclassification rate.
\begin{theorem}
	\label{thm:downstream} If Assumption~\ref{ap:downstream} holds, and $m$ and $K$ are upper bounded by some constant, then, with probability $1-C_{m,K}\chi_n$, we have
	$$
	\xi(\hat{\Theta})\le C_K \chi_n.
	$$
	Here, $C_{m,K}$ is some constant relying on $m$ and $K$ and $C_K$ is some constant relying on $K$.
\end{theorem}

Theorem~\ref{thm:downstream} demonstrates that the performance of the downstream linear classifier hinges on the quality of our representation in approximating a linearly separable one, rather than the sample size in the downstream analysis. This result implies that a linear classifier, learned from a limited number of labeled samples, can achieve a low misclassification rate when self-supervised learning effectively captures an accurate linearly separable representation through a large quantity of unlabeled samples. By combining Theorem~\ref{thm:downstream} with the findings from preceding sections, we can further compare the impacts of supervised and self-supervised learning on the downstream linear classifier. Let $\hat{\Theta}_{L}$ and $\hat{\Theta}_{\bar{L}}$ be the data representations acquired through the classical Laplacian-based method and augmentation invariant manifold learning, respectively. When $(\log n/n)^{1/d_s}\ll \delta(\Mcal)\ll (\log n/n)^{1/d}$, with a probability tending to 1, 
$$
\lim_{n\to \infty}\xi(\hat{\Theta}_{\bar{L}})= 0,
$$
and there exists an instance (corresponding to the scenario in Theorem~\ref{thm:lowerml}) such that
$$
\lim_{n\to \infty}\xi(\hat{\Theta}_{L})\ge 1/2.
$$

\section{Numerical Illustration}
\label{sc:numerical}
In this section, we present a series of numerical experiments to validate the theoretical results from the previous sections and to compare the performance of unsupervised and self-supervised learning methods. Specifically, we utilize the MNIST dataset \citep{lecun1998gradient}, which consists of 60,000 training and 10,000 testing images of $28\times28$ gray-scale handwritten digits. For our self-supervised learning approach, we adopt the transformation introduced by \cite{wang2022augmentation} to generate augmented data. This involves random resizing, cropping, and rotation of the images.

In all the numerical experiments, we compare the performance of Augmentation Invariant Manifold Learning (AIML) as defined in \eqref{eq:aimlop}, with that of a continuous version of the classical graph Laplacian-based method (CML). To comprehensively study the impact of data augmentation, we formulate the unsupervised graph Laplacian-based method as a similar optimization problem to \eqref{eq:aimlop}, but with the removal of all components related to data augmentation:
\begin{equation}
	\label{eq:cml_op}
	\min_{\beta\in\Bcal}\quad \sum_{i,j=1}^nW_{i,j}\|\Theta_\beta(X_{i})-\Theta_\beta(X_{j})\|^2+\lambda_2\Rcal(\Theta_\beta),
\end{equation}
where $W_{i,j}$ represents the weights between $X_i$ and $X_{j}$, and $\Theta_\beta$ is the encoder. This optimization problem can be regarded as a computationally efficient and continuous version of the classical graph Laplacian-based method. For a fair comparison, we employ the same convolutional neural network encoder with two convolution+ReLU layers, two pooling layers, and a fully connected layer. Additionally, we use identical tuning parameters and optimization algorithms for both AIML and CML.

\begin{figure}[t!]
	\centering
	\includegraphics[width=0.45\textwidth]{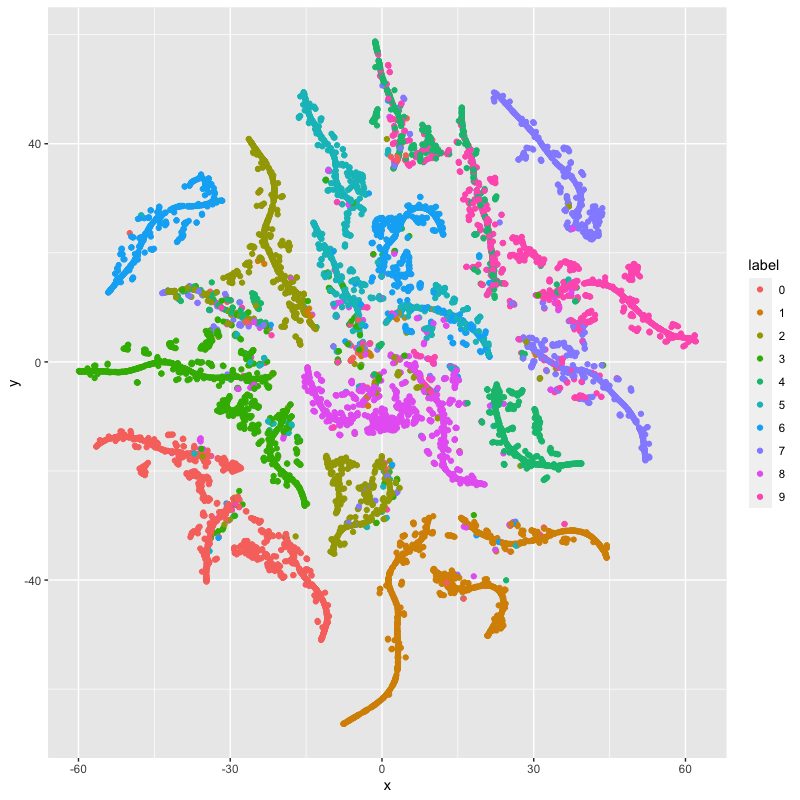}
	\hskip 20pt
	\includegraphics[width=0.45\textwidth]{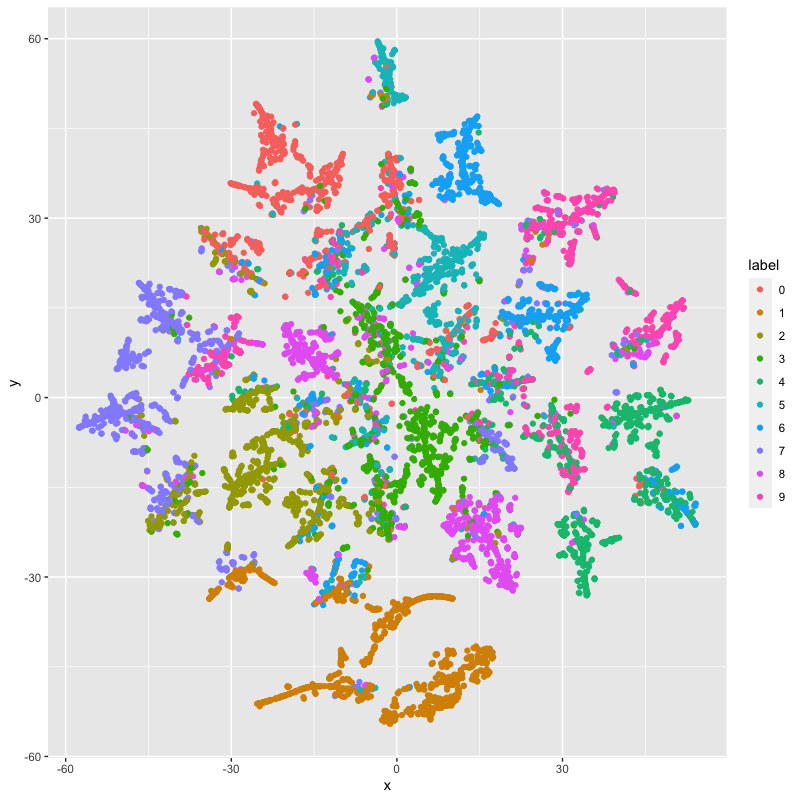}
	\caption{t-SNE plots of representation learned by augmentation invariant manifold learning (left) and graph Laplacian-based method (right).}
	\label{fg:tsne}
\end{figure}

We present t-SNE plots \citep{hinton2002stochastic} depicting the learned representations in Figure~\ref{fg:tsne} when applying these two methods to the training images. It is important to note that t-SNE plots can distort the distances between data points, but they offer a visual representation of the cluster structure in the learned representations. As shown in Figure~\ref{fg:tsne}, each digit corresponds to multiple clusters, and leveraging data augmentation enhances the separation of distinct digits.

\begin{figure}[t!]
	\centering
	\includegraphics[width=0.4\textwidth]{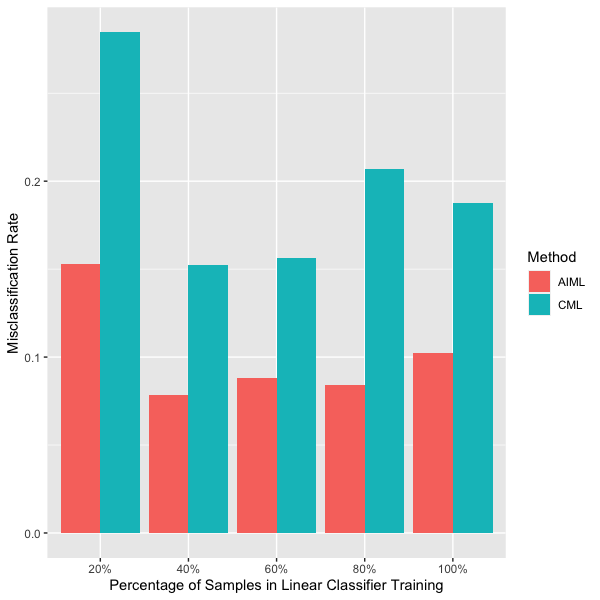}
	\hskip 20pt
	\includegraphics[width=0.4\textwidth]{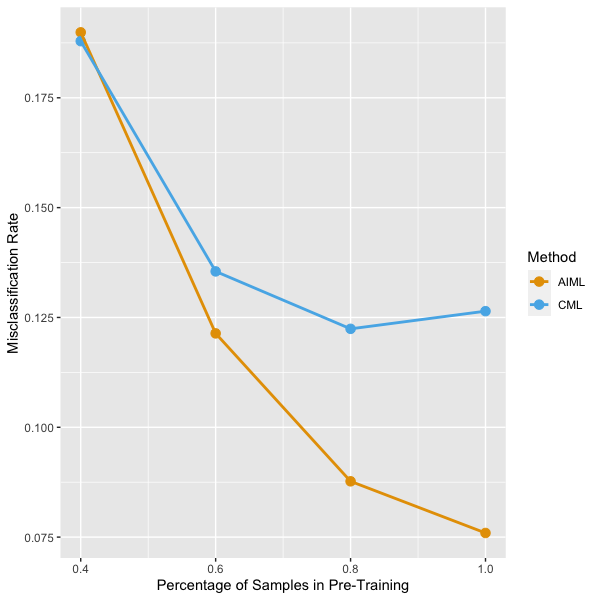}
	\caption{Misclassification rate of AIML and CML: the left figure shows the result when the sample size in representation learning is fixed and in downstream task varies; the right figure shows the result when the sample size in representation learning varies and in downstream task is fixed.}
	\label{fg:accuracy}
\end{figure}

To quantitatively assess linear separability, we train a linear classifier on top of the learned representations and measure the resulting classifier's accuracy. We recode the digit labels as $1$ when the digit is smaller than $5$ and as $-1$ when the digit is equal to or larger than $5$. Thus, the classification task aims to predict whether a digit is smaller than $5$. We conduct two sets of numerical experiments to evaluate the influence of sample size on representation learning (via unsupervised or self-supervised methods) and classifier training in the downstream task. In the first set of experiments, we use all training samples (without labels) to learn data representations, and the linear classifier is trained using 20\%, 40\%, 60\%, 80\%, and 100\% of labeled samples. The left figure in Figure~\ref{fg:accuracy} presents the misclassification rates for AIML and CML representations. It's evident that the performance of AIML and CML remains stable as the sample size increases from 40\% to 100\%, confirming the findings in Theorem~\ref{thm:downstream}. In the second set of experiments, we learn the representation from 40\%, 60\%, 80\%, and 100\% of unlabeled training samples, and the downstream classifier is trained with 40\% of labeled samples. The resulting linear classifier's misclassification rates are reported in the right figure of Figure~\ref{fg:accuracy}. Clearly, a larger sample size in representation learning leads to a more accurate classifier. Combining the outcomes of both experiment sets, we observe that data augmentation aids in learning better linearly separable representations, and the classifier's accuracy primarily depends on the separability of learned representations, rather than the downstream analysis's sample size. These experimental results align with the theoretical findings presented in previous sections.

\section{Concluding Remarks}
\label{sc:conclusion}

In this paper, we delve into the significance of data augmentation in learning linearly separable representations. Our investigation underscores the pivotal role played by the invariant structures introduced through data augmentation, facilitating the better transformation of intricate nonlinear data structures into linearly separable representations via unlabeled datasets. Coupled with the insights gleaned from the findings in \cite{wang2022augmentation}, we conclude that data augmentation yields two principal advantages within self-supervised representation learning: enhanced separation of closely situated manifolds and nonlinear dimension reduction within each distinct manifold. Furthermore, exploring the implications for downstream analysis reveals that the crux of effective classification lies in the successful segregation of data originating from distinct manifolds. Remarkably, even with a limited number of labeled samples, the availability of an efficient linearly separable representation empowers the construction of accurate classifiers. This observation elucidates the feasibility of developing efficient algorithms in the context of few-shot learning \citep{wang2020generalizing}, particularly in scenarios where an extensive unlabeled dataset coexists with a smaller labeled counterpart.

\section*{Acknowledgment}
This project is supported by grants from the National Science Foundation (DMS-2113458).

\bibliographystyle{plainnat}
\bibliography{MultiManifoldRepresentation}

\begin{appendices}
\section{Algorithm for Augmentation Invariant Manifold Learning}

This section presents the stochastic optimization algorithm for  augmentation invariant manifold learning, summarized in Algorithm~\ref{ag:aimlop}.

\begin{algorithm}[h!]
	\caption{Augmentation Invariant Manifold Learning}\label{ag:aimlop}
	\begin{algorithmic}[1]
		\REQUIRE A set of data $\{X_1,\ldots, X_n\}$, batch size $n'$, encoder $\Theta_\beta$, stochastic data augmentation transformation $\Tcal$, tuning parameters $(r,\lambda_1,\lambda_2)$.
		\FOR{sampled minibatch $\{X_i: i\in \Scal \}$} 
		\STATE Generate two independent augmented copies of each sample $X'_i=\Tcal(X_i)$ and $X''_i=\Tcal(X_i)$ for $i\in \Scal$.
		\STATE Evaluate representation of each augmented sample $Z'=\{\Theta_\beta(X'_i)\}_{i\in \Scal}$ and $Z''=\{\Theta_\beta(X''_i)\}_{i\in \Scal}$, where $Z',Z''\in \RR^{n'\times S}$.
		\STATE Evaluate the kernel matrix $W=\{\bI(\|X'_i-X''_{j}\|\le r)\}_{i,j\in\Scal}\in \RR^{n'\times n'}$ and corresponding Laplacian matrix $L$.
		\STATE Evaluate the loss 
		$$
		\hat{\ell}(\beta)={\rm tr}({Z'}^TLZ'')+\lambda_1\|Z'-Z''\|_F^2+\lambda_2\|{Z'}^TZ''-I_S\|_F^2,
		$$
		where $I_S$ is a $S\times S$ identify matrix and $\|\cdot\|_F$ is Frobenius norm of a matrix.
		\STATE Update $\Theta_\beta$ to minimize $\hat{\ell}(\beta)$.
		\ENDFOR
		\ENSURE Encoder $\Theta_\beta$
	\end{algorithmic}
\end{algorithm}

\section{Proof}
In this proof, $\|\cdot\|$ represents the Euclidean distance and $d_\Mcal(\cdot,\cdot)$ represents the geodesic distance on the manifold $\Mcal$. $C$ refers to be a constant which can be different in different places.

\subsection{Proof for Theorem~\ref{thm:upperml}}

Instead of proving Theorem~\ref{thm:upperml} directly, we prove a more general version of Theorem~\ref{thm:upperml}. 
\begin{theorem}
	Suppose Assumption~\ref{ap:ml} holds and for any $1\le k\le K$
	$$
	\pi_k\left(B(x,r)\bigcap \Mcal_k\right)\le \beta,\qquad x\in \Mcal\setminus\Mcal_k,
	$$
	where $B(x,r)$ is a ball centered at $x$ with radius $r$ (in Euclidean distance). If we assume $r\to 0$,
	$$
	r\gg \left(\log n\over n\right)^{1/d}\qquad {\rm and}\qquad {\beta\over r^{d+2}}+{\sqrt{\beta}\over nr^{d+2}}\to 0,
	$$
	then with probability at least $1-K^2/t^2-4Kn^{-\alpha}$ for some $\alpha>0$, we have 
	$$
	|\lambda_s(L)-\lambda_s(\Mcal)|\le e_1\lambda_2(\Mcal)+e_2 ,
	$$
	where $\lambda_s(L)$ is $s$th eigenvalue of normalized matrix $L$ in $b(U)$ and $\lambda_s(\Mcal)$ is $s$th eigenvalue of $\Delta_{\Mcal}$. Here, $e_1$ and $e_2$ are defined as
	$$
	e_1=C\left(r \sqrt{\lambda_s(\Mcal)}+\gamma+r+{\delta\over r}\right)\quad {\rm and}\quad e_2={C\left(\beta n +t\sqrt{\beta}+t\beta\sqrt{n}\over nr^{d+2}\right)}\left(1+\lambda_s^{d/2+2}(\Mcal)\right).
	$$
	Here, we choose $\delta=\sqrt{r(c\log n/n)^{1/d}}$ and $\gamma=\sqrt{(\alpha+1)\log n/n\delta^d}$ for some large enough $\alpha$ so $\delta,\gamma, \delta/r \to 0$. 
	With probability at least $1-K^2/t^2-4Kn^{-\alpha}$, if $U_s$ is normalized eigenvector of $L$ with eigenvalue $\lambda_s(L)$, there is a normalized eigenfunction $\theta_s$ of $\Delta_{\Mcal}$ with eigenvalue $\lambda_s(\Mcal)$ such that
	$$
	\|U_s-\vec{\theta}_s\|_{L^2(\pi_n)}\le \begin{cases} C\left(\gamma+\delta+{1\over \sqrt{\gamma_\lambda}} \left(\beta n +t\sqrt{\beta}+t\beta\sqrt{n}\over nr^{d+2}\right)^{1/2}\right), & s\le K\\ C\left({\lambda \over \gamma_\lambda}\left( r \sqrt{\lambda}+\gamma+r+{\delta\over r}\right)+{\left(\beta n +t\sqrt{\beta}+t\beta\sqrt{n}\over nr^{d+2}\right)}\lambda^{d/2+2}\right)^{1/2}+C\delta, &s>K \end{cases},
	$$
	where $\lambda=\lambda_s(\Mcal)$, $\vec{\theta}_s=(\theta_s(X_1),\ldots, \theta_s(X_n))$ and $\gamma_\lambda$ is the eigengap. 
\end{theorem}
This theorem can lead to Theorem~\ref{thm:upperml} if we choose $\beta=0$. This theorem also suggests that the first $K$ eigenvectors converge faster than the rest of eigenvectors.

\paragraph{Step 1: Dirichlet energy} To study the convergence of graph Laplacian, we need to introduce some notations. For a function $\theta:\Mcal\to \RR$ defined on $\Mcal$, we also write it in the following form
$$
\theta(x)=(\theta_1(x),\ldots,\theta_K(x)),
$$
where each $\theta_k:\Mcal_k\to \RR$ is a function defined on each manifold $\Mcal_k$. Given two functions $\theta^A$ and $\theta^B$ defined on $\Mcal$, we define their inner product as
$$
\langle \theta^A,\theta^B\rangle_{L^2(\pi)} =\sum_{k=1}^K w_k\int_{\Mcal_k}\theta_k^A(x)\theta^B_k(x) \pi_k(x)dx.
$$
We also define the weighted Dirichlet energy for a function $\theta$ as
$$
D(\theta)=\sum_{k=1}^K w_k^2 D_k(\theta_k),\qquad {\rm where} \quad D_k(\theta_k)=\int_{\Mcal_k} \|\nabla \theta_k(x)\|^2\pi_k^2(x)dx.
$$
From this definition, it is straightforward to see $D(\theta)=\langle \theta,\Delta_{\Mcal}\theta\rangle$. 

\paragraph{Step 2: Discrete Dirichlet energy} To study graph Laplacian, we now introduce a more general neighborhood graph and the corresponding graph Laplacian. Specifically, when $\|X_i-X_j\|\le r$, we assign weights 
$$
W_{i,j}=h\left(\|X_i-X_j\|\over r\right).
$$
where $h$ is a function supported on $[0,1]$. Clearly, $h(x)=\bI(\|x\|\le 1)$ used in Section~\ref{sc:manifoldlearning} is a special case. Given the general weights, we can define discrete Dirichlet energy as
$$
b(U)={1\over  n^2}\sum_{i,j}{1\over \sigma_h r^d}W_{i,j}\left(U(X_i)-U(X_j)\over r\right)^2={1 \over \sigma_h n^2r^{d+2} }U^TLU,
$$
where $U(X_i)$ is the $i$th component of vector $U$ (corresponds to $X_i$) and $\sigma_h$ is the surface tension of $h$
$$
\sigma_h=\int |y_1|^2h(|y|)dy.
$$
In particular, when $h(x)=\bI(\|x\|\le 1)$, $\sigma_h=V_d/(d+2)$, where $V_d$ is the volume of $d$-dimensional Euclidean unit ball. Although $U\in \RR^n$, it can also be considered as a function defined on discrete point $\{X_1,\ldots,X_n\}$. If we write the empirical measure of $\{X_1,\ldots,X_n\}$ as
$$
\pi_n={1\over n}\sum_{i=1}^n \bI_{X_i},
$$
then we can define the inner product between two discrete functions
$$
\langle U^A,U^B\rangle_{L^2(\pi_n)} ={1\over n}\sum_{i=1}^n U^A(X_i)U^B(X_i).
$$
Similarly, we can define within and cross manifold Dirichlet energy
$$
b_k(U)={1\over n_k^2}\sum_{X_i,X_j\in \Mcal_k}{1\over \sigma_h r^d}W_{i,j}\left(U(X_i)-U(X_j)\over r\right)^2,\qquad b_W(U)=\sum_{k=1}^K \left(n_k\over n\right)^2b_k(U),
$$
and 
$$
b_C(U)={1\over n^2} \sum_{k_1\ne k_2}\sum_{X_i\in \Mcal_{k_1}, X_j\in \Mcal_{k_2}}{1\over \sigma_h r^d}W_{i,j}\left(U(X_i)-U(X_j)\over r\right)^2.
$$
Here, $n_k$ is the number of point on $\Mcal_k$, i.e., $n_k=|\{i:X_i\in \Mcal_k\}|$. 

\paragraph{Step 3: Discretization and interpolation maps} After introducing the notations, we now construct the relationship between Dirichlet energy $D(\theta)$ and $b(U)$ via discretization and interpolation maps. The Proposition 2.12 in \cite{calder2022improved} or Corollary A.3 in \cite{trillos2021large} suggests the following Proposition directly. We omit the proof.
\begin{proposition}
	\label{prop:maps}
	With probability at least $1-Kn\exp(-Cn\gamma^2\delta^d)-2K\exp(-cn)$, there exists a probability measure $\tilde{\mu}_{n,k}$ with probability density function $\tilde{\pi}_{n,k}$ for $k=1,\ldots, K$ such that
	$$
	\|\pi_k-\tilde{\pi}_{n,k}\|_{L^\infty(\Mcal_k)}\le C(\gamma+\delta)
	$$
	and there exist  transportation maps $\tilde{R}_1,\ldots, \tilde{R}_K$ such that
	$$
	\sup_{x\in \Mcal_k}d_{\Mcal_k}(x,\tilde{R}_k(x))\le \delta.
	$$
\end{proposition}
In Proposition~\ref{prop:maps}, we can choose $\delta=\sqrt{r(c\log n/n)^{1/d}}$ and $\gamma=\sqrt{(\alpha+1)\log n/n\delta^d}$ for some large enough $\alpha$.
Based on the transportation maps $\tilde{R}_1,\ldots, \tilde{R}_K$ in Proposition~\ref{prop:maps}, we can also introduce the discretization and interpolation maps. First, we can define a partition of $\Mcal$ by $\tilde{R}_1,\ldots, \tilde{R}_K$. Specifically, if $X_i\in \Mcal_k$, then 
$$
\tilde{U}_{i}=\tilde{R}_k^{-1}(X_{i})\subset \Mcal_k.
$$
After defining the partition, we can introduce the discretization map $\tilde{P}: L^2(\pi) \to L^2(\pi_n)$
$$
\tilde{P}\theta(X_{i})=n_k\int_{\tilde{U}_{i}}\theta(x)\tilde{\pi}_{n,k}(x)dx,
$$
where $X_i\in \Mcal_k$. Similarly, we can introduce the associated extension map $\tilde{P}^\ast: L^2(\pi_n) \to L^2(\pi)$
$$
\tilde{P}^\ast U(x)=\sum_{i=1}^n u(X_{i})\bI(x\in \tilde{U}_{i}).
$$
We then define the interpolation map $\tilde{\Ical}$
$$
\tilde{\Ical}u=\Lambda_{r-2\delta} \tilde{P}^\ast u
$$
where $\Lambda_{r-2\delta}$ is a convolution operator defined on $\Mcal$. More concretely, we define
$$
\Lambda_{r}\theta(x)={\int K_r(x,y)\theta_k(y)dy\over \int K_r(x,y)dy},\qquad x\in \Mcal_k,
$$
where the kernel $K_r(x,y)$ is defined as
$$
K_r(x,y)={1\over r^d}\psi\left(d_\Mcal(x,y)\over r\right)\qquad {\rm and}\qquad \psi(t)={1\over \sigma_\eta}\int_t^\infty h(s)ds.
$$

\paragraph{Step 4: Connection between Dirichlet energy $D(\theta)$ and $b(U)$} With the newly introduced discretization and interpolation maps, we can connect the Dirichlet energy $D(\theta)$ and $b(U)$. We can directly apply the results from Proposition 4.1 and 4.2 in \cite{calder2022improved} or Proposition A.4 and A.5 in \cite{trillos2021large} to show the following Proposition. 
\begin{proposition}
	\label{prop:within}
	 Suppose $\gamma$ and $\delta$ are two parameters in Proposition~\ref{prop:maps}. With probability at least $1-Kn\exp(-Cn\gamma^2\delta^d)-2K\exp(-cn)$, we have
	\begin{equation}
		\label{eq:dirconnect}
		b_W(\tilde{P}\theta)\le \left(1+C\left({\delta\over r}+r+\gamma\right)\right)D(\theta)\quad {\rm and}\quad D(\tilde{\Ical}U)\le \left(1+C\left({\delta\over r}+r+\gamma\right)\right)b(U). 
	\end{equation}
	In addition, we also have 
	\begin{equation}
		\label{eq:discretediff}
		\left|\|\theta\|^2_{L^2(\pi)}-\|\tilde{P}\theta\|^2_{L^2(\pi_n)}\right|\le C\delta \|\theta\|_{L^2(\pi)}\sqrt{D(\theta)}+C(\gamma+\delta)\|\theta\|^2_{L^2(\pi)}
	\end{equation}
	and
	\begin{equation}
		\label{eq:continuediff}
		\left|\|U\|^2_{L^2(\pi_n)}-\|\tilde{\Ical}U\|^2_{L^2(\pi)}\right|\le Cr \|U\|_{L^2(\pi_n)}\sqrt{b(U)}+C(\gamma+\delta)\|U\|^2_{L^2(\pi_n)}.
	\end{equation}
Here, $\theta\in L^2(\pi)$ and $U\in L^2(\pi_n)$.
\end{proposition}
To connect Dirichlet energy $D(\theta)$ and $b(U)$, we also need to find an upper bound for the cross manifold Dirichlet energy. 
\begin{proposition}
	\label{prop:cross}
	Suppose Assumption~\ref{ap:ml} holds and for any $1\le k\le K$
	$$
	\pi_k\left(B(x,r)\bigcap \Mcal_k\right)\le \beta,\qquad \forall\ x\in \Mcal\setminus\Mcal_k,
	$$
	where $B(x,r)$ is a ball centered at $x$ with radius $r$ (in Euclidean distance). When $\theta$ is in the span of $\Delta_{\Mcal}$'s eigenfunctions with corresponding eigenvalue less than $\lambda$, then we have 
	$$
	b_C(\tilde{P}\theta)\le {C\left(\beta n +t\sqrt{\beta}+t\beta\sqrt{n}\over nr^{d+2}\right)}\left(1+\lambda^{d/2+2}\right)\|\theta\|^2_{L^2(\pi)},
	$$
	with probability at least $1-K^2/t^2-2K\exp(-cn)$.
\end{proposition}

\paragraph{Step 5: Upper bound of $\lambda_s(L)$} To show the convergence of eigenvalues, we now construct an upper bound for $\lambda_s(L)$ in terms of $\lambda_s(\Mcal)$. Let $\theta^1,\ldots,\theta^s$ be an orthonormal set of eigenfunctions of $\Delta_{\Mcal}$. Then, we can define 
$$
V^j=\tilde{P}\theta^j,\qquad j=1,\ldots, s.
$$
We can apply \eqref{eq:discretediff} to $\theta^{j_1}-\theta^{j_2}$, so we have 
\begin{align*}
	&\left|\|\theta^{j_1}\|^2_{L^2(\pi)}+\|\theta^{j_2}\|^2_{L^2(\pi)}-2\langle \theta^{j_1},\theta^{j_2}\rangle_{L^2(\pi)}-\|V^{j_1}\|^2_{L^2(\pi_n)}-\|V^{j_2}\|^2_{L^2(\pi_n)}+2\langle V^{j_1},V^{j_2}\rangle_{L^2(\pi_n)}\right|\\
	\le & C\delta \sqrt{\lambda_s(\Mcal)}+C(\gamma+\delta)
\end{align*}
So we can conclude that for some large enough $n$, we have 
$$
\left|\langle \theta^{j_1},\theta^{j_2}\rangle_{L^2(\pi)}-\langle V^{j_1},V^{j_2}\rangle_{L^2(\pi_n)}\right|\le C\delta \sqrt{\lambda_s(\Mcal)}+C(\gamma+\delta)<{1\over s}.
$$
Because $\langle \theta^{j_1},\theta^{j_2}\rangle_{L^2(\pi)}=1$ when $j_1=j_2$ and $\langle \theta^{j_1},\theta^{j_2}\rangle=0$ when $j_1\ne j_2$, we can know $V^1,\ldots, V^s$ are linearly independent. Based on $V^1,\ldots, V^s$, we define a $s$-dimensional subspace
$$
S^o=\left\{\sum_{j=1}^sa_jV^j:a_j\in \RR\right\}.
$$
By the Courant-Fisher minimax principle, we know
$$
\lambda_s(L)=\min_{S\in \Scal_s}\max_{U\in S\setminus \{0\}}{b(U)\over \|U\|^2_{L^2(\pi_n)}},
$$
where $\Scal_s$ is the collection of all possible $s$-dimensional subspaces. This immediately suggests
$$
\lambda_s(L)\le \max_{U\in S^o, \|U\|_{L^2(\pi_n)}=1} b(U).
$$
For any $U\in S^o$, we can apply \eqref{eq:dirconnect} and Proposition~\ref{prop:cross} to obtain
\begin{align*}
	b(U)=b(\tilde{P}\theta)&\le \left(1+C\left({\delta\over r}+r+\gamma\right)\right)D(\theta)+{C\left(\beta n +t\sqrt{\beta}+t\beta\sqrt{n}\over nr^{d+2}\right)}\left(1+\lambda^{d/2+2}\right)\|\theta\|^2_{L^2(\pi)}\\
	&\le \left(1+C\left({\delta\over r}+r+\gamma\right)\right)\lambda_s(\Mcal) \|\theta\|^2_{L^2(\pi)}+{C\left(\beta n +t\sqrt{\beta}+t\beta\sqrt{n}\over nr^{d+2}\right)}\left(1+\lambda^{d/2+2}\right)\|\theta\|^2_{L^2(\pi)}.
\end{align*}
where $\theta=\sum_{j=1}^sa_j\theta^j$ if $U=\sum_{j=1}^sa_jV^j$ and $\lambda=\lambda_s(\Mcal)$.
An application of \eqref{eq:discretediff} suggests
$$
\|\theta\|^2_{L^2(\pi)}\le \|U\|^2_{L^2(\pi_n)}+ C(\delta \sqrt{\lambda_s(\Mcal)}+\gamma+\delta) \|\theta\|^2_{L^2(\pi)}.
$$
If $\|U\|_{L^2(\pi_n)}=1$, then
$$
\|\theta\|^2_{L^2(\pi)}\le 1+C(\delta \sqrt{\lambda_s(\Mcal)}+\gamma+\delta).
$$
This leads to
$$
\lambda_s(L)\le \left(1+C\left(\delta \sqrt{\lambda_s(\Mcal)}+\gamma+r+{\delta\over r}\right)\right)\lambda_s(\Mcal)+{C\left(\beta n +t\sqrt{\beta}+t\beta\sqrt{n}\over nr^{d+2}\right)}\left(1+\lambda_s^{d/2+2}(\Mcal)\right).
$$

\paragraph{Step 6: Lower bound of $\lambda_s(L)$} We can find a lower bound for $\lambda_s(L)$ in terms of $\lambda_s(\Mcal)$ with a similar idea. Let $U^1,\ldots,U^s$ be an orthonormal set of eigenvectors of $L$ and
$$
\theta^j=\tilde{\Ical}U^j,\qquad j=1,\ldots, s.
$$
We can apply \eqref{eq:continuediff} to $U^{j_1}-U^{j_2}$
\begin{align*}
	&\left|\|\theta^{j_1}\|^2_{L^2(\pi)}+\|\theta^{j_2}\|^2_{L^2(\pi)}-2\langle \theta^{j_1},\theta^{j_2}\rangle_{L^2(\pi)}-\|U^{j_1}\|^2_{L^2(\pi_n)}-\|U^{j_2}\|^2_{L^2(\pi_n)}+2\langle U^{j_1}, U^{j_2}\rangle_{L^2(\pi_n)}\right|\\
	\le & Cr \sqrt{\lambda_s(L)}+C(\gamma+\delta)
\end{align*}
So we can conclude that for some large enough $n$, we have 
$$
\left|\langle \theta^{j_1},\theta^{j_2}\rangle_{L^2(\pi)}-\langle U^{j_1},U^{j_2}\rangle_{L^2(\pi_n)}\right|\le Cr \sqrt{\lambda_s(L)}+C(\gamma+\delta)<{1\over s}.
$$
Because $\langle U^{j_1},U^{j_2}\rangle_{L^2(\pi_n)}=1$ when $j_1=j_2$ and $\langle U^{j_1},U^{j_2}\rangle_{L^2(\pi_n)}=0$ when $j_1\ne j_2$, we can know $\theta^1,\ldots, \theta^s$ are linearly independent. With $\theta^1,\ldots, \theta^s$, we define 
$$
S^o=\left\{\sum_{j=1}^sa_j\theta^j:a_j\in \RR\right\}.
$$
Again, by the Courant-Fisher minimax principle, 
$$
\lambda_s(\Mcal)=\min_{S\in \Scal_s}\max_{\theta\in S\setminus \{0\}}{D(\theta)\over \|\theta\|^2_{L^2(\pi)}}\le \max_{\theta\in S^o, \|\theta\|_{L^2(\pi)}=1} D(\theta).
$$
For $\theta\in S^o$, there exist a $U$ such that $\theta=\tilde{\Ical}U$ where $U=\sum_{j=1}^sa_jU^j$. If we apply \eqref{eq:dirconnect}, then we have 
$$
D(\theta)=b(\tilde{\Ical}U)\le \left(1+C\left({\delta\over r}+r+\gamma\right)\right)b(U)\le \left(1+C\left({\delta\over r}+r+\gamma\right)\right)\lambda_s(L) \|U\|^2_{L^2(\pi_n)}.
$$
An application of \eqref{eq:dirconnect} again suggests that if $\|\theta\|_{L^2(\pi)}=1$, then 
$$
\|U\|^2_{L^2(\pi_n)}\le 1+C(r \sqrt{\lambda_s(L)}+\gamma+\delta).
$$
Now we can conclude
$$
\lambda_s(\Mcal)\le \left(1+C\left(r \sqrt{\lambda_s(L)}+\gamma+r+{\delta\over r}\right)\right)\lambda_s(L).
$$
Putting upper and lower bounds together yields
$$
|\lambda_s(\Mcal)-\lambda_s(L)| \le C\left(r \sqrt{\lambda_s(\Mcal)}+\gamma+r+{\delta\over r}\right)\lambda_s(\Mcal)+{C\left(\beta n +t\sqrt{\beta}+t\beta\sqrt{n}\over nr^{d+2}\right)}\left(1+\lambda_s^{d/2+2}(\Mcal)\right).
$$
In the rest of proof, we use the following notations
$$
e_1=C\left(r \sqrt{\lambda_s(\Mcal)}+\gamma+r+{\delta\over r}\right)\qquad {\rm and}\qquad e_2={C\left(\beta n +t\sqrt{\beta}+t\beta\sqrt{n}\over nr^{d+2}\right)}\left(1+\lambda_s^{d/2+2}(\Mcal)\right).
$$

\paragraph{Step 7: Convergence of the first $K$ eigenvectors}
After showing the convergence of eigenvalues, we now study the convergence of  eigenvectors. We first show the convergence of the eigenvectors corresponding to the first $K$ eigenvalues ($\lambda_1(\Mcal)=\ldots=\lambda_K(\Mcal)=0$). Let $\gamma_\lambda$ be the eigengap, i.e., $\gamma_\lambda=\lambda_{K+1}(\Mcal)$. When $n$ is large enough, we have
$$
|\lambda_s(\Mcal)-\lambda_s(L)|\le {\gamma_\lambda\over 4}, \qquad {\rm when}\quad s=1,\ldots, K+1.
$$
Let $S$ be the subspace spanned by the eigenvectors of $L$ with eigenvalues $\lambda_1(L),\ldots,\lambda_K(L)$ and $\Pcal_S$ be the projection operator on $S$. Let $\theta$ be a normalized eigenvector of $\Delta_{\Mcal}$ with eigenvalue $0$ and $U=\tilde{P}\theta$. By Proposition~\ref{prop:within} and \ref{prop:cross}, we can know that
$$
b(U)\le {C\left(\beta n +t\sqrt{\beta}+t\beta\sqrt{n}\over nr^{d+2}\right)}.
$$ 
The definition of eigenvector suggests
$$
b(U)={1 \over \sigma_h n^2r^{d+2} }U^TLU\ge \lambda_1(L)\|\Pcal_SU\|^2_{L^2(\pi_n)}+\lambda_{K+1}(L)\|U-\Pcal_SU\|^2_{L^2(\pi_n)}=\lambda_{K+1}(L)\|U-\Pcal_SU\|^2_{L^2(\pi_n)}.
$$
Putting above two inequalities together yields
\begin{equation}
	\label{eq:eigenprojection}
	\|U-\Pcal_SU\|^2_{L^2(\pi_n)}\le {C\over \gamma_\lambda} {\left(\beta n +t\sqrt{\beta}+t\beta\sqrt{n}\over nr^{d+2}\right)}.
\end{equation}
By the definition of $\tilde{P}$, we can know that if $X_i\in \Mcal_k$, then
$$
|U(X_{i})-\theta(X_{i})|=\left|n_k\int_{\tilde{U}_{i}}\left(\theta(x)-\theta(X_{i})\right)\tilde{\pi}_{n,k}(x)dx\right|\le \|\nabla \theta_k\|_\infty \delta.
$$
Since $\theta$ is the eigenvector of $\Delta_{\Mcal}$ with eigenvalue $0$, then we can know $\|\nabla \theta_k\|_\infty=0$ and $U=\vec{\theta}$, where $\vec{\theta}=(\theta(X_1),\ldots,\theta(X_n))\in \RR^n$ is a $n$ dimensional vector. So we can know  \begin{equation}
	\label{eq:projection}
	\|\vec{\theta}-\Pcal_S\tilde{P}\theta\|^2_{L^2(\pi_n)}\le {C\over \gamma_\lambda} {\left(\beta n +t\sqrt{\beta}+t\beta\sqrt{n}\over nr^{d+2}\right)}.
\end{equation}
Let $\theta^1,\ldots,\theta^K$ be a set of orthonormal basis for the eigenspace of eigenfunctions of $\Delta_{\Mcal}$ with eigenvalue $0$, $\tilde{U}^j=\tilde{P}\theta^j$ and $\tilde{V}^j=\Pcal_S\tilde{P}\theta^j$. By \eqref{eq:projection}, we have 
$$
\|\vec{\theta}^j-\tilde{V}^j\|^2_{L^2(\pi_n)}=\|\tilde{U}^j-\tilde{V}^j\|^2_{L^2(\pi_n)}\le {C\over \gamma_\lambda} {\left(\beta n +t\sqrt{\beta}+t\beta\sqrt{n}\over nr^{d+2}\right)}.
$$
An application of \eqref{eq:discretediff} on $\theta^{j_1}-\theta^{j_2}$ suggests
$$
\left|\langle \tilde{U}^{j_1},\tilde{U}^{j_2}\rangle_{L^2(\pi_n)}-\langle \theta^{j_1},\theta^{j_2}\rangle_{L^2(\pi)} \right|\le C(\gamma+\delta).
$$
We combine \eqref{eq:eigenprojection} and above inequality to obtain
$$
\left|\langle \tilde{V}^{j_1},\tilde{V}^{j_2}\rangle_{L^2(\pi_n)}-\langle \theta^{j_1},\theta^{j_2}\rangle_{L^2(\pi)}\right|\le C\left(\gamma+\delta+{1\over \sqrt{\gamma_\lambda}} \left(\beta n +t\sqrt{\beta}+t\beta\sqrt{n}\over nr^{d+2}\right)^{1/2}\right).
$$
Let $V^{1},\ldots,V^{K}$ be the Gram-Schmidt orthogonalization of $\tilde{V}^{1},\ldots, \tilde{V}^{K}$. Then we can know that 
$$
\|\tilde{V}^{j}-V^j\|_{L^2(\pi_n)}\le  C\left(\gamma+\delta+{1\over \sqrt{\gamma_\lambda}} \left(\beta n +t\sqrt{\beta}+t\beta\sqrt{n}\over nr^{d+2}\right)^{1/2}\right).
$$
Therefore, we can conclude that
$$
\|V^{j}-\vec{\theta}^j\|_{L^2(\pi_n)}\le  C\left(\gamma+\delta+{1\over \sqrt{\gamma_\lambda}} \left(\beta n +t\sqrt{\beta}+t\beta\sqrt{n}\over nr^{d+2}\right)^{1/2}\right).
$$
Equivalently, if we apply some rotation matrix, we can find an orthonormal set $\theta'_1,\ldots,\theta'_K$ of eigenfunctions of $\Delta_{\Mcal}$ with eigenvalues $0$ such that
$$
\|U_{j}-\vec{\theta}'_j\|_{L^2(\pi_n)}\le  C\left(\gamma+\delta+{1\over \sqrt{\gamma_\lambda}} \left(\beta n +t\sqrt{\beta}+t\beta\sqrt{n}\over nr^{d+2}\right)^{1/2}\right).
$$

\paragraph{Step 8: Convergence of the rest eigenvectors} We next study the convergence of the other eigenvectors by induction. In particular, let $\lambda_{s-1}(\Mcal)<\lambda=\lambda_s(\Mcal)=\ldots=\lambda_{s+l-1}(\Mcal)<\lambda_{s+l}(\Mcal)$. Suppose $\theta^1,\ldots,\theta^{s-1}$ are a set of orthonormal basis for the eigenspace of eigenfunctions of $\Delta_{\Mcal}$ with eigenvalue smaller than $\lambda$ and there exists a set of orthonormal basis for subspace spanned by the eigenvectors of $L$ with eigenvalues smaller than $\lambda$ such that
\begin{equation}
	\label{eq:induction}
	\|V^j-\tilde{P}\theta^j\|_{L^2(\pi_n)}\le C\left({1\over \gamma_\lambda}\left( r \sqrt{\lambda}+\gamma+r+{\delta\over r}\right)+{C\left(\beta n +t\sqrt{\beta}+t\beta\sqrt{n}\over nr^{d+2}\right)}\left(1+\lambda^{d/2+2}\right)\right)^{1/2}+C\delta
\end{equation}
for $j=1,\ldots,s-1$. Let $S$ be the subspace spanned by the eigenvectors of $L$ with eigenvalues $\lambda_s(L),\ldots,\lambda_{s+l-1}(L)$ and $\Pcal_S$ be the projection operator on $S$. Similarly, we define $S_+$ as the subspace spanned by the eigenvectors of $L$ with eigenvalues larger than $\lambda_{s+l-1}(L)$ and $S_-$ as the subspace spanned by the eigenvectors of $L$ with eigenvalues smaller than $\lambda_{s}(L)$. $\Pcal_{S_+}$ and $\Pcal_{S_-}$ are the projection operator on $S_+$ and $S_-$. Let $\theta$ be an eigenvector of $\Delta_{\Mcal}$ with eigenvalue $\lambda$ and let $U=\tilde{P}\theta$. By Proposition~\ref{prop:within} and \ref{prop:cross}, we have
$$
b(U)\le {C\left(\beta n +t\sqrt{\beta}+t\beta\sqrt{n}\over nr^{d+2}\right)}\left(1+\lambda^{d/2+2}\right)+\left(1+C\left({\delta\over r}+r+\gamma\right)\right)\lambda.
$$
By the spectrum decomposition, we have
\begin{align*}
	b(U)&={1 \over \sigma_h n^2r^{d+2} }U^TLU \\
	& \ge \lambda_s(L)\|\Pcal_SU\|^2_{L^2(\pi_n)}+\lambda_{s+l}(L)\|\Pcal_{S_+}U\|^2_{L^2(\pi_n)}\\
	& \ge \lambda_s(L)\left(\|U\|^2_{L^2(\pi_n)}-\|U-\Pcal_SU\|^2_{L^2(\pi_n)}\right)+\lambda_{s+l}(L)\left(\|U-\Pcal_SU\|^2_{L^2(\pi_n)}-\|\Pcal_{S_-}U\|^2_{L^2(\pi_n)}\right)\\
	&\ge \lambda_s(L)\|U\|^2_{L^2(\pi_n)}+(\lambda_{s+l}(L)-\lambda_s(L))\|U-\Pcal_SU\|^2_{L^2(\pi_n)}-\lambda_{s+l}(L)\|\Pcal_{S_-}U\|^2_{L^2(\pi_n)}.
\end{align*}
By \eqref{eq:discretediff}, we can know 
$$
\left|\|U\|^2_{L^2(\pi_n)}-1\right|\le C\delta \sqrt{\lambda}+C(\gamma+\delta).
$$
The results in step 6 suggests 
$$
|\lambda-\lambda_k(L)| \le C\left(r \sqrt{\lambda}+\gamma+r+{\delta\over r}\right)\lambda+{C\left(\beta n +t\sqrt{\beta}+t\beta\sqrt{n}\over nr^{d+2}\right)}\left(1+\lambda^{d/2+2}\right).
$$
Therefore, we can have
$$
{\gamma_\lambda\over 2}\|U-\Pcal_SU\|^2_{L^2(\pi_n)}\le C\left(r \sqrt{\lambda}+\gamma+r+{\delta\over r}\right)\lambda+{C\left(\beta n +t\sqrt{\beta}+t\beta\sqrt{n}\over nr^{d+2}\right)}\lambda^{d/2+2}+\lambda_{k+l}(L)\|\Pcal_{S_-}U\|^2_{L^2(\pi_n)}.
$$
By the choice of orthonormal basis $V^1,\ldots,V^{s-1}$, we have 
\begin{align*}
	\Pcal_{S_-}u=\sum_{j=1}^{s-1}\langle V^j, U\rangle_{L^2(\pi_n)} V^j.
\end{align*}
Since \eqref{eq:induction} and
$$
\left|\langle \tilde{P}\theta^{j}, U\rangle_{L^2(\pi_n)}\right|\le C\delta \sqrt{\lambda}+C(\gamma+\delta),
$$ 
which is implied by \eqref{eq:discretediff}, we can know 
$$
\left|\langle V^j, U\rangle_{L^2(\pi_n)}\right|\le C\delta \sqrt{\lambda}+C\left({1\over \gamma_\lambda}\left( r \sqrt{\lambda}+\gamma+r+{\delta\over r}\right)+{C\left(\beta n +t\sqrt{\beta}+t\beta\sqrt{n}\over nr^{d+2}\right)}\lambda^{d/2+2}\right)^{1/2}.
$$
This leads to 
$$
\|\Pcal_{S_-}U\|_{L^2(\pi_n)}\le C\delta \sqrt{\lambda}+C\left({1\over \gamma_\lambda}\left( r \sqrt{\lambda}+\gamma+r+{\delta\over r}\right)+{C\left(\beta n +t\sqrt{\beta}+t\beta\sqrt{n}\over nr^{d+2}\right)}\lambda^{d/2+2}\right)^{1/2}.
$$
The convergence of eigenvalue suggests
$$
|\lambda_{s+l}(\Mcal)-\lambda_{s+l}(L)| \le C\left(r \sqrt{\lambda_{k+l}(\Mcal)}+\gamma+r+{\delta\over r}\right)\lambda_{s+l}(\Mcal)+{C\left(\beta n +t\sqrt{\beta}+t\beta\sqrt{n}\over nr^{d+2}\right)}\lambda_{s+l}^{d/2+2}(\Mcal).
$$
This immediately leads to 
$$
{\gamma_\lambda\over 2}\|U-\Pcal_SU\|^2_{L^2(\pi_n)}\le C\left(r \sqrt{\lambda}+\gamma+r+{\delta\over r}\right)\lambda+{C\left(\beta n +t\sqrt{\beta}+t\beta\sqrt{n}\over nr^{d+2}\right)}\lambda^{d/2+2}+C\delta^2.
$$
Therefore, we can conclude that
$$
\|U-\Pcal_SU\|^2_{L^2(\pi_n)}\le C\left(r \sqrt{\lambda}+\gamma+r+{\delta\over r}\right){\lambda\over \gamma_\lambda}+ {C\left(\beta n +t\sqrt{\beta}+t\beta\sqrt{n}\over nr^{d+2}\right)}\lambda^{d/2+2}+C\delta^2.
$$
By the definition of $\tilde{P}$, we can know that if $X_i\in \Mcal_k$, then
$$
|U(X_{i})-\theta(X_{i})|=\left|n_k\int_{\tilde{U}_{i}}\left(\theta(x)-\theta(X_{i})\right)\tilde{\pi}_{n,k}(x)dx\right|\le \|\nabla \theta_k\|_\infty \delta.
$$
Since $\theta$ is the eigenvector of $\Delta_{\Mcal}$, then we can know $\|\nabla \theta_k\|_\infty$ is finite.
So we can know  
$$
	\|\vec{\theta}-\Pcal_S\tilde{P}\theta\|^2_{L^2(\pi_n)}\le C\left(r \sqrt{\lambda}+\gamma+r+{\delta\over r}\right){\lambda\over \gamma_\lambda}+ {C\left(\beta n +t\sqrt{\beta}+t\beta\sqrt{n}\over nr^{d+2}\right)}\lambda^{d/2+2}+C\delta^2.
$$
Then, we can follow the same argument in step 7 to show that if $U_1,\ldots, U_{s+l-1}$ is an orthonormal basis for the eigenspace of $L$ with eigenvalues smaller than $\lambda_{s+l}(L)$, then there exists an orthonormal basis for the eigenspace of $\Delta_{\Mcal}$ with eigenvalues smaller than $\lambda_{s+l}(\Mcal)$ such that
$$
\|U_{j}-\vec{\theta}_j\|_{L^2(\pi_n)}\le  C\left({\lambda \over \gamma_\lambda}\left( r \sqrt{\lambda}+\gamma+r+{\delta\over r}\right)+{\left(\beta n +t\sqrt{\beta}+t\beta\sqrt{n}\over nr^{d+2}\right)}\lambda^{d/2+2}\right)^{1/2}+C\delta.
$$

\subsection{Proof for Theorem~\ref{thm:lowerml}}
In this proof, we adopt the same notations in proof of Theorem~\ref{thm:upperml}. Besides the Dirichlet energy $D(\theta)$ and $b(U)$, we define a new discrete Dirichlet energy based on $Y_1,\ldots, Y_n$
$$
b_{Y}(U)={1\over  n^2}\sum_{i,j}{1\over \sigma_h \tilde{r}^d}\tilde{W}_{i,j}\left(U(Y_i)-U(Y_j)\over \tilde{r}\right)^2,
$$
where $\tilde{r}=\sqrt{r^2-{z^o}^2}$ and the weight $\tilde{W}_{i,j}$ is defined by $Y_i$ and $Y_j$
$$
\tilde{W}_{i,j}=\bI\left(\|Y_i-Y_j\|\le \tilde{r}\right).
$$
We also define a Dirichlet energy on $\tilde{\Mcal}$
$$
D_Y(\theta)=\int_{\tilde{\Mcal}}\|\nabla \theta(y)\|\pi_Y(y)dy,
$$
where $\pi_Y$ is a uniform distribution on $\tilde{\Mcal}$. Similarly, we can also introduce the discretization map $\tilde{P}_Y$ and interpolation map $\tilde{\Ical}_Y$ by $Y_1,\ldots, Y_n$ on $\tilde{\Mcal}$. We choose $\delta=\sqrt{r(c\log n/n)^{1/d}}$ and $\gamma=\sqrt{(\alpha+1)\log n/n\delta^d}$ for some large enough $\alpha$ in $\tilde{P}_Y$ and $\tilde{\Ical}_Y$. Instead of finding the connection between $D(\theta)$ and $b(U)$, we will build the connection between $D_Y(\theta)$ and $b(U)$. 

By the construction of $\Mcal_1$ and $\Mcal_2$, we have
$$
\|Y_i-Y_j\|\le \tilde{r}\qquad \Rightarrow \qquad \|X_i-X_j\|\le r.
$$
Equivalently, we have 
$$
\bI\left(\|Y_i-Y_j\|\le \tilde{r}\right)\le \bI\left(\|X_i-X_j\|\le r\right)\qquad {\rm or}\qquad \tilde{W}_{i,j}\le W_{i,j}.
$$
This suggests
\begin{align*}
	b_{Y}(U)&={1\over  n^2}\sum_{i,j}{1\over \sigma_h \tilde{r}^d}\tilde{W}_{i,j}\left(U(Y_i)-U(Y_j)\over \tilde{r}\right)^2\\
	&\le {1\over  n^2}\sum_{i,j}{1\over \sigma_h \tilde{r}^d}W_{i,j}\left(U(Y_i)-U(Y_j)\over \tilde{r}\right)^2\\
	&\le \left(r\over \tilde{r}\right)^{d+2}b(U).
\end{align*}
By Proposition~\ref{prop:within}, we have 
\begin{align*}
D_Y(\tilde{\Ical}_YU)&\le \left(1+C\left({\delta\over \tilde{r}}+\tilde{r}+\gamma\right)\right)b_Y(U) \\
&\le \left(1+C\left({\delta\over \tilde{r}}+\tilde{r}+\gamma\right)\right) \left(r\over \tilde{r}\right)^{d+2}b(U)\\
&\le \left(1+C\left({\delta\over \tilde{r}}+\tilde{r}+\gamma+\left(z^o\over \tilde{r}\right)^2\right)\right)b(U).
\end{align*}

On the other hand, if we write $U=\tilde{P}_Y\theta$ where $\theta$ is some eigenfucntion of $\Delta_{\tilde{\Mcal}}$, then 
\begin{align*}
	&\left(r \over \tilde{r}\right)^{d+2}b(U)-b_Y(U)\\
	=&{1\over \sigma_h n^2\tilde{r}^{d+2}}\sum_{Z_i=Z_j} \left(\bI(\|Y_i-Y_j\|\le r)-\bI(\|Y_i-Y_j\|\le \tilde{r})\right)(U(Y_i)-U(Y_j))^2\\
	=&{1\over \sigma_h n^2\tilde{r}^{d}}\sum_{Z_i=Z_j} \bI(\tilde{r}<\|Y_i-Y_j\|\le r)\left(U(Y_i)-U(Y_j) \over \tilde{r}\right)^2\\
	\le& {1\over \sigma_h n^2\tilde{r}^{d}}\sum_{Z_i=Z_j}\bI(\tilde{r}<\|Y_i-Y_j\|\le r)\left(3r\|\nabla\theta\|_\infty\over \tilde{r}\right)^2\\
	\le & {9r^2\|\nabla\theta\|_\infty^2\over \sigma_h n^2\tilde{r}^{d+2}}\sum_{i,j}\bI(\tilde{r}<\|Y_i-Y_j\|\le r)
\end{align*}
Define a $U$-statistics
$$
Q={2\over n(n-1)}\sum_{i<j}\bI(\tilde{r}<\|Y_i-Y_j\|\le r). 
$$
Since 
$$
\Sigma^2:=\EE\left(\bI^2(\tilde{r}<\|Y_i-Y_j\|\le r)\right)=\EE\left(\bI(\tilde{r}<\|Y_i-Y_j\|\le r)\right)\le C(r^d-\tilde{r}^d),
$$
we can apply a Bernstein-type inequality for $U$-statistics \citep{arcones1995bernstein} to have
$$
\PP\left(\sqrt{n}|Q-\EE(Q)|>t\right)\le 2\exp\left(-{2t^2\over 2\Sigma^2+(2/3)tn^{-1/2} }\right).
$$
This suggest
$$
\PP\left(Q>C(r^d-\tilde{r}^d)+\sqrt{C(r^d-\tilde{r}^d)\log n\over n}+{C\log n\over n}\right)\le 2n^{-2}
$$
With probability at least $1-2n^{-2}$, 
$$
\left(r \over \tilde{r}\right)^{d+2}b(U)-b_Y(U)\le {Cr^2\|\nabla\theta\|_\infty^2\over \sigma_h \tilde{r}^{d+2}}\left((r^d-\tilde{r}^d)+{\log n\over n}\right)\le {C\|\nabla\theta\|_\infty^2\over \sigma_h}\left({{z^o}^2\over r^2}+{\log n\over nr^d}\right).
$$
Therefore, we can conclude that
\begin{align*}
b(\tilde{P}_Y\theta)&\le b_Y(\tilde{P}_Y\theta)+ {C\|\nabla\theta\|_\infty^2\over \sigma_h}\left({{z^o}^2\over r^2}+{\log n\over nr^d}\right)\\
&\le \left(1+C\left({\delta\over \tilde{r}}+\tilde{r}+\gamma\right)\right)D_Y(\theta)+ {C\|\nabla\theta\|_\infty^2\over \sigma_h}\left({{z^o}^2\over r^2}+{\log n\over nr^d}\right).
\end{align*}
After building the connection between $D_Y(\theta)$ and $b(U)$, we can apply the same arguments in step 5-8 of proof for Theorem~\ref{thm:upperml} to show that, with probability at least $1-Cn^{-2}$, if $U_s$ is normalized eigenvector of $L$ with $s$th eigenvalue, there is a normalized eigenfunction $\theta_s$ of $\Delta_{\tilde{\Mcal}}$ with $s$th eigenvalue such that
$$
\|U_s-\vec{\theta}_s\|_{L^2(\pi_n)}\le C\left({\lambda \over \gamma_\lambda}\left( r \sqrt{\lambda}+\gamma+r+{\delta\over r}+\left(z^o\over \tilde{r}\right)^2\right)+{C\over \sigma_h}\left(\left(z^o\over r\right)^2+{\log n\over nr^d}\right)\right)^{1/2}+C\delta, 
$$
where $\lambda=\lambda_s(\tilde{\Mcal})$, $\vec{\theta}_s=(\theta_s(Y_1),\ldots, \theta_s(Y_n))$ and $\gamma_\lambda$ is the eigengap. The choices of $\delta$, $\gamma$, $z^o$ and $r$ suggest
$$
\|U_s-\vec{\theta}_s\|_{L^2(\pi_n)}\to 0.
$$

\subsection{Proof for Theorem~\ref{thm:upperaiml}}

We adopt a similar strategy in Theorem~\ref{thm:upperml} to prove results. 
\paragraph{Step 1: Dirichlet energy} We need to introduce some notations parallel to the notations in the proof of Theorem~\ref{thm:upperml}. Given two functions $\bar{\theta}^A$ and $\bar{\theta}^B$ defined on $\Ncal_s$, we define their inner product as
$$
\langle \bar{\theta}^A,\bar{\theta}^B\rangle_\phi =\sum_{k=1}^K w_k\int_{\Ncal_{s,k}}\bar{\theta}_k^A(\phi)\bar{\theta}^B_k(\phi) \pi^s_k(\phi)d\phi.
$$
Similarly, we define the weighted Dirichlet energy for a function $\bar{\theta}:\Ncal_s\to \RR$ as
$$
D_\phi(\bar{\theta})=\sum_{k=1}^K {w_k^2\over {\rm Vol}\Ncal_{v,k}} D_{\phi,k}(\bar{\theta}_k),\qquad {\rm where} \quad D_{\phi,k}(\bar{\theta}_k)=\int_{\Ncal_{s,k}} \|\nabla \bar{\theta}_k(\phi)\|^2{\pi_k^s}^2(\phi)d\phi.
$$

\paragraph{Step 2: Discrete Dirichlet energy} 
Given the new weights, we can define corresponding discrete Dirichlet energy as
$$
b_\phi(\bar{U})={1\over  n^2}\sum_{i,j}{1\over \sigma_h r^{d}}\bar{W}_{i,j}\left(\bar{U}(\phi_i)-\bar{U}(\phi_j)\over r\right)^2,
$$
where $\bar{U}(\phi_i)$ is the $i$th component of vector $\bar{U}$ and $\sigma_h$ is the surface tension when 
$$
h(x)=(1-x^2)_+^{d_v/2}\qquad {\rm where}\quad (y)_+=\max(y,0). 
$$
We can define within and cross manifold Dirichlet energy
$$
b_{\phi,k}(\bar{U})={1\over n_k^2}\sum_{\phi_i,\phi_j\in \Ncal_{s,k}}{1\over \sigma_h r^{d}}\bar{W}_{i,j}\left(\bar{U}(\phi_i)-\bar{U}(\phi_j)\over r\right)^2,\qquad b_{\phi,W}(\bar{U})=\sum_{k=1}^K \left(n_k\over n\right)^2b_{\phi,k}(\bar{U}),
$$
and 
$$
b_{\phi,C}(\bar{U})={1\over n^2} \sum_{k_1\ne k_2}\sum_{\phi_i\in \Ncal_{s,k_1}, \phi_j\in \Ncal_{s,k_2}}{1\over \sigma_h r^{d}}\bar{W}_{i,j}\left(\bar{U}(\phi_i)-\bar{U}(\phi_j)\over r\right)^2.
$$
We also need to define an intermediate discrete Dirichlet energy
$$
\tilde{b}_\phi(\bar{U})={1\over  n^2}\sum_{i,j}{1\over \sigma_h r^{d_s}}h\left(d_{\Ncal_{s,k}}(\phi_i,\phi_j)\over r\right)\left(\bar{U}(\phi_i)-\bar{U}(\phi_j)\over r\right)^2,
$$
and corresponding within manifold Dirichlet energy
$$
\tilde{b}_{\phi,W}(\bar{U})=\sum_{k=1}^K \left(n_k\over n\right)^2{1\over {\rm Vol}\Ncal_{v,k}}b_{\phi,k}(\bar{U}),
$$
where 
$$
\tilde{b}_{\phi,k}(\bar{U})={1\over n_k^2}\sum_{\phi_i,\phi_j\in \Ncal_{s,k}}{1\over \sigma_h r^{d_s}}h\left(d_{\Ncal_{s,k}}(\phi_i,\phi_j)\over r\right)\left(\bar{U}(\phi_i)-\bar{U}(\phi_j)\over r\right)^2.
$$

\paragraph{Step 3: Discretization and interpolation maps} By applying Proposition~\ref{prop:maps}, we can know that with probability at least $1-Kn\exp(-Cn\gamma^2\delta^{d_s})-2K\exp(-cn)$, there exists a probability measure $\tilde{\mu}^s_{n,k}$ with probability density function $\tilde{\pi}^s_{n,k}$ for $k=1,\ldots, K$ such that
$$
\|\pi^s_k-\tilde{\pi}^s_{n,k}\|_{L^\infty(\Ncal_{s,k})}\le C(\gamma+\delta)
$$
and there exist  transportation maps $\tilde{R}_1,\ldots, \tilde{R}_K$ such that
$$
\sup_{x\in \Ncal_{s,k}}d_{\Ncal_{s,k}}(x,\tilde{R}_k(x))\le \delta.
$$
Similarly, we can choose $\delta=\sqrt{r(c\log n/n)^{1/d_s}}$ and $\gamma=\sqrt{(\alpha+1)\log n/n\delta^{d_s}}$ for some large enough $\alpha$. Based on the transportation maps $\tilde{R}_1,\ldots, \tilde{R}_K$, we can introduce the discretization map $\tilde{P}_\phi: L^2(\pi^s) \to \RR^n$ and the interpolation map $\tilde{\Ical}_\phi: \RR^n \to L^2(\pi^s)$ in the same way as the proof of Theorem~\ref{thm:upperml}.

\paragraph{Step 4: Connection between Dirichlet energy $D_\phi(\bar{\theta})$ and $b_\phi(\bar{U})$} To build the connection between $D_\phi(\bar{\theta})$ and $b_\phi(\bar{U})$, we need to find more explicit upper and lower bound for $\bar{W}_{i,j}$. If $\phi_i,\phi_j\in \Ncal_{s,k}$ and $d_{\Ncal_{s,k}}(\phi_i,\phi_j)\le r$, then
\begin{align*}
	{1\over r^{d}}\bar{W}_{i,j}&={1\over r^{d}{\rm Vol}^2\Ncal_{v,k}}\int_{\Mcal(\phi_i)}\int_{\Mcal(\phi_j)}\bI(\|x-y\|\le r)dxdy\\
	&\ge {1\over r^{d}{\rm Vol}^2\Ncal_{v,k}}\int_{\Mcal(\phi_i)}\int_{\Mcal(\phi_j)}\bI\left(d_{\Mcal_k}(x,y)\le r\right)dxdy\\
	&\ge {1\over r^{d}{\rm Vol}^2\Ncal_{v,k}}\int_{\Ncal_{v,k}}\int_{\Ncal_{v,k}}\bI\left(d_{\Ncal_{v,k}}(\psi_i,\psi_j)\le \sqrt{r^2-d^2_{\Ncal_{s,k}}(\phi_i,\phi_j)}\right)d\psi_id\psi_j\\
	&\ge {1\over r^d{\rm Vol}^2\Ncal_{v,k}}\int_{\Ncal_{v,k}}\left(1-C(r^2-d^2_{\Ncal_{s,k}}(\phi_i,\phi_j))\right)V_{d_v}\left(r^2-d^2_{\Ncal_{s,k}}(\phi_i,\phi_j)\right)^{d_v/2}  d\psi_1\\
	&\ge {V_{d_v}\over r^d{\rm Vol}\Ncal_{v,k}}\left(1-C(r^2-d^2_{\Ncal_{s,k}}(\phi_i,\phi_j))\right)\left(r^2-d^2_{\Ncal_{s,k}}(\phi_i,\phi_j)\right)^{d_v/2}\\
	&\ge {V_{d_v}\over r^{d_s}{\rm Vol}\Ncal_{v,k}}\left(1-Cr^2\right)\left(1-{d^2_{\Ncal_{s,k}}(\phi_i,\phi_j)\over r^2}\right)^{d_v/2}.
\end{align*}
Here we use the following facts
\begin{enumerate}
	\item $\|x-y\|\le d_{\Mcal_k}(x,y)$;
	\item $d^2_{\Mcal_k}(x,y)=d^2_{\Ncal_{s,k}}(\phi_i,\phi_j)+d^2_{\Ncal_{v,k}}(\psi_i,\psi_j)$ when $x=T_k(\phi_i,\psi_i)$ and $y=T_k(\phi_j,\psi_j)$;
	\item $|{\rm Vol}(B_{\Ncal_{v,k}}(\psi,r))-V_{d_v}r^{d_v}|\le Cr^{d_v+2}$, where $B_{\Ncal_{v,k}}(\psi,r)$ is a geodesic ball with center at $\psi$ and radius $r$;
	\item $\Ncal_{v,k}$ has no boundary.
\end{enumerate}
This suggests
$$
{1\over r^{d}}\bar{W}_{i,j}\ge {V_{d_v}\over r^{d_s}{\rm Vol}\Ncal_{v,k}}\left(1-Cr^2\right)h\left(d_{\Ncal_{s,k}}(\phi_i,\phi_j)\over r\right)
$$
and 
$$
b_{\phi,k}(\bar{U})\ge \left(1-Cr^2\right){V_{d_v}\over {\rm Vol}\Ncal_{v,k}} \tilde{b}_{\phi,k}(\bar{U}).
$$
If we apply Proposition~\ref{prop:within}, we have
\begin{align*}
b_{\phi,W}(\bar{U})&\ge \left(1-Cr^2\right)V_{d_v}\tilde{b}_{\phi,W}(\bar{U})\\
&\ge \left(1-Cr^2\right)V_{d_v} D_{\phi}(\tilde{\Ical}_\phi\bar{U})\big/\left(1+C\left({\delta\over r}+r+\gamma\right)\right).
\end{align*}
Because $b_{\phi,C}(\bar{U})\ge 0$, 
\begin{align*}
D_{\phi}(\tilde{\Ical}_\phi\bar{U})&\le {1\over V_{d_v}}\left(1+C\left({\delta\over r}+r+\gamma+r^2\right)\right)b_{\phi,W}(\bar{U})\\
&\le {1\over V_{d_v}}\left(1+C\left({\delta\over r}+r+\gamma+r^2\right)\right)b_{\phi}(\bar{U}).
\end{align*}

Next, we work on the upper bound of $\bar{W}_{i,j}$.  If $\phi_i,\phi_j\in \Ncal_{s,k}$, then
\begin{align*}
	{1\over r^d}\bar{W}_{i,j}&={1\over r^d{\rm Vol}^2\Ncal_{v,k}}\int_{\Mcal(\phi_i)}\int_{\Mcal(\phi_j)}\bI(\|x-y\|\le r)dxdy\\
	&\le {1\over r^d{\rm Vol}^2\Ncal_{v,k}}\int_{\Mcal(\phi_i)}\int_{\Mcal(\phi_j)}\bI\left(d_{\Mcal}(x,y)\le \tilde{r}\right)dxdy\\
	&\le {1\over r^d{\rm Vol}^2\Ncal_{v,k}}\int_{\Ncal_{v,k}}\int_{\Ncal_{v,k}}\bI\left(d_{\Ncal_{v,k}}(\psi_i,\psi_j)\le \sqrt{\tilde{r}^2-d^2_{\Ncal_{s,k}}(\phi_i,\phi_j)}\right)d\psi_id\psi_j\\
	&\le {1\over r^d{\rm Vol}^2\Ncal_{v,k}}\int_{\Ncal_{v,k}}\left(1+c\tilde{r}^2\right)V_{d_v}(\tilde{r}^2-d^2_{\Ncal_{s,k}}(\phi_i,\phi_j))^{d_v/2}  d\psi_i\\
	&\le {1\over r^d{\rm Vol}\Ncal_{v,k}}\left(1+c\tilde{r}^2\right)V_{d_v}\left(\tilde{r}^2-d^2_{\Ncal_{s,k}}(\phi_i,\phi_j)\right)^{d_v/2}\\
	&\le {(\tilde{r}/r)^{d_v}\over r^{d_s}{\rm Vol}\Ncal_{v,k}}\left(1+c\tilde{r}^2\right)V_{d_v}\left(1-{d^2_{\Ncal_{s,k}}(\phi_i,\phi_j)\over \tilde{r}^2}\right)^{d_v/2}
\end{align*}
Here $\tilde{r}=r+8r^3/R^2$ and we use the following facts
\begin{enumerate}
	\item $d_{\Mcal_k}(x,y)\le \|x-y\|+8\|x-y\|^3/R^2$, where $R$ is the reach of the manifold;
	\item $d^2_{\Mcal_k}(x,y)=d^2_{\Ncal_{s,k}}(\phi_i,\phi_j)+d^2_{\Ncal_{v,k}}(\psi_i,\psi_j)$ when $x=T_k(\phi_i,\psi_i)$ and $y=T_k(\phi_j,\psi_j)$;
	\item $|{\rm Vol}(B_{\Ncal_{v,k}}(\psi,r))-V_{d_v}r^{d_v}|\le Cr^{d_v+2}$, where $B_{\Ncal_{v,k}}(\psi,r)$ is a geodesic ball with center at $\psi$ and radius $r$;
	\item $\Ncal_{v,k}$ has no boundary.
\end{enumerate}
So we can have 
$$
{1\over r^{d}}\bar{W}_{i,j}\le \left(\tilde{r}\over r\right)^d{V_{d_v}\over \tilde{r}^{d_s}{\rm Vol}\Ncal_{v,k}}\left(1+C\tilde{r}^2\right)h\left(d_{\Ncal_{s,k}}(\phi_i,\phi_j)\over \tilde{r}\right)
$$
and 
$$
b_{\phi,k}(\bar{U})\le \left(\tilde{r}\over r\right)^d\left(1+Cr^2\right){V_{d_v}\over {\rm Vol}\Ncal_{v,k}} \tilde{b}_{\phi,k}(\bar{U}).
$$
Note that the size of neighborhood in $\tilde{b}_{\phi,k}(\bar{U})$ of the above inequality is $\tilde{r}$. An application of Proposition~\ref{prop:within} leads to 
\begin{align*}
	b_{\phi,W}(\tilde{P}_\phi\bar{\theta})&\le \left(1+Cr^2\right)V_{d_v}\tilde{b}_{\phi,W}(\tilde{P}_\phi\bar{\theta})\\
	&\le \left(1+Cr^2\right)\left(1+C\left({\delta\over \tilde{r}}+\tilde{r}+\gamma\right)\right)V_{d_v} D_{\phi}(\bar{\theta})\\
	&\le \left(1+C\left({\delta\over \tilde{r}}+\tilde{r}+\gamma+r^2\right)\right)V_{d_v} D_{\phi}(\bar{\theta}).
\end{align*}
When $\delta(\Mcal)>r$, we can know $b_{\phi,C}(\bar{U})=0$. This immediately suggests
$$
b_{\phi}(\tilde{P}_\phi\bar{\theta})\le \left(1+C\left({\delta\over \tilde{r}}+\tilde{r}+\gamma+r^2\right)\right)V_{d_v} D_{\phi}(\bar{\theta}).
$$

After building the connection between $D_\phi(\bar{\theta})$ and $b_\phi(\bar{U})$, we can apply the same arguments in step 5-8 of proof for Theorem~\ref{thm:upperml} to complete the proof.

\subsection{Proof for Theorem~\ref{thm:lowergeneral}}
The proof is divided into three steps.
	
	\paragraph{Step 1: hypothesis construction} In this step, we construct the least favorable hypothesis $\HH_0$ and $\HH_1$.  Under $\HH_0$, we assume the observed data is drawn from a single manifold
	$$
	\Mcal=\Mcal_0=\{(x,0_{D-d}):x\in[0,1]^d\},
	$$
	where $0_{D-d}$ is a $D-d$ dimensional vector of zeros. We also assume that the probability distribution of our observed data is uniform distribution on $\Mcal_0$. Given the manifold, we write $\PP_0$ as the probability distribution of observed data $X_1,\ldots,X_n$ under $\HH_0$. Next, we construct the other hypothesis $\HH_1$. We  divide the space $[0,1]^d$ into $L=M^d$ cubes of size $M^{-1}\times \ldots \times M^{-1}$ for some positive integer $M$ and name these disjoint cubes $Q_1,\ldots, Q_L$. $M$ will be specified in the third step. For each $Q_l$, we also define $\tilde{Q}_l$ as a cube with the same center of $Q_l$ but a smaller size  $(3M)^{-1}\times \ldots \times (3M)^{-1}$. Given $Q_l$, we consider the following two manifolds
	$$
	\Mcal_1(Q_l)=\{(x,0_{D-d}):x\in[0,1]^d\setminus Q_l\}\qquad {\rm and}\qquad \Mcal_2(Q_l)=\{(x,0_{D-d}):x\in\tilde{Q}_l\}.
	$$
	Clearly, $\Mcal_1(Q_l)$ and $\Mcal_2(Q_l)$ are two disjoint manifolds and the distance between them is $(3M)^{-1}$. So $\delta(\Mcal(Q_l))=(3M)^{-1}$ if we define $\Mcal(Q_l)=\Mcal_1(Q_l) \cup \Mcal_2(Q_l)$. Given these two manifolds, we consider a uniform distribution over $\Mcal(Q_l)$ and define it as hypothesis $\HH_1(Q_l)$. Write  $\PP_1(Q_l)$ as the probability distribution of observed data $X_1,\ldots,X_n$ under $\HH_1(Q_l)$. The hypothesis $\HH_1$ is a mixture of hypotheses $\HH_1(Q_l)$ for $l=1, \ldots, L$ and we write 
	$$
	\PP_1={1\over L}\sum_{l=1}^L\PP_1(Q_l).
	$$
	By the construction, the observed data is drawn from single manifold under $\HH_0$ and two disjoint manifolds under $\HH_1$.
	
	\paragraph{Step 2: bounding $\chi^2$ divergence} The goal of this step is to bound $\chi^2$ divergence between $\PP_0$ and $\PP_1$. To the end, we write the likelihood between $\PP_1(Q_l)$ and $\PP_0$ as 
	$$
	F_l(X_1,\ldots X_n)={d\PP_1(Q_l) \over d\PP_0}=\begin{cases}
		\left({1\over 1-\Delta}\right)^n,&\qquad \forall X_i\in ([0,1]^d\setminus Q_l )\cup \tilde{Q}_l\\
		0, &\qquad \exists X_i\in Q_l \setminus \tilde{Q}_l
	\end{cases},
	$$
	where $\Delta=M^{-d}-(3M)^{-d}$ is the volume of set $Q_l\setminus\tilde{Q}_l$. Given the likelihood $F_1,\ldots, F_L$, we can write the $\chi^2$ divergence between $\PP_0$ and $\PP_1$ as
	$$
	\chi^2(\PP_1,\PP_0)=\EE_0\left({d\PP_1\over d\PP_0}-1\right)^2=\Var_0\left({1\over L}\sum_{l=1}^LF_l\right),
	$$
	where $\EE_0$ and $\Var_0$ are the expectation and variance under the probability distribution $\PP_0$. Therefore, it is sufficient to bound the variance and covariance of $F_1,\ldots, F_L$. For any $l_1\ne l_2$, we have 
	$$
	F_{l_1}F_{l_2}=\begin{cases}
		\left({1\over 1-\Delta}\right)^{2n},&\qquad \forall X_i\in \left([0,1]^d\setminus (Q_{l_1}\cup Q_{l_2}) \right)\cup \tilde{Q}_{l_1}\cup \tilde{Q}_{l_2}\\
		0, &\qquad \exists X_i\in (Q_{l_1} \setminus \tilde{Q}_{l_1})\cup (Q_{l_2} \setminus \tilde{Q}_{l_2})
	\end{cases}.
	$$
	Because $\EE_0(F_l)=1$, we can bound the covariance of $F_{l_1}$ and $F_{l_2}$ in the following way
	\begin{align*}
		\left|\Cov_0(F_{l_1},F_{l_2})\right|&=\left|\EE_0(F_{l_1}F_{l_2})-1\right|\\
		&=\left|\left({1\over 1-\Delta}\right)^{2n}\left(1-2\Delta\right)^n-1\right|\\
		&=\left|\left(1-{\Delta^2\over (1-\Delta)^2}\right)^n-1\right|\\
		&\le 1-\exp\left(-n\Delta^2/(1-\Delta)\sqrt{1-2\Delta}\right)\\
		&\le {n\Delta^2\over (1-\Delta)\sqrt{1-2\Delta}}.
	\end{align*}
	Here we the fact that $\log(1-x)\ge -x/\sqrt{1-x}$ when $0<x<1$ and $1-x\le e^{-x}$. For the variance term, we have 
	\begin{align*}
		\Var_0(F_{l})&\le \EE_0(F_l^2)= \left({1\over 1-\Delta}\right)^{n}\\
		&= \exp \left(n\log \left(1+{\Delta\over 1-\Delta}\right)\right)\\
		&\le \exp \left({n\Delta\over 1-\Delta}\right),
	\end{align*}
	where we use the fact $\log(1+x)\le x$. Putting the bound of covariance and variance together yields
	$$
	\chi^2(\PP_1,\PP_0)\le {1\over L}\exp \left({n\Delta\over 1-\Delta}\right)+{n\Delta^2\over (1-\Delta)\sqrt{1-2\Delta}}.
	$$
	
	\paragraph{Step 3: wrap up the proof} Let $T$ be any test. We still need to specify the choice of $M$ in the hypotheses constructed in step 1. Specifically, we can choose $M=\lceil (2n/\log n)^{1/d} \rceil$, so 
	$$
	L\ge 2n/\log n\qquad {\rm and}\qquad \Delta\le {\log n\over 2n}.
	$$
	Therefore, we have 
	$$
	\chi^2(\PP_1,\PP_0)\le {\log n \over n^{1/3}}+{\log^2 n\over n}\to 0.
	$$
	Therefore, we have 
	\begin{align*}
		\PP_0(T=1)+\PP_1(T=0)\ge 1-\chi^2(\PP_1,\PP_0)\to 1.
	\end{align*}
	The construction suggests that under the hypothesis $\HH_1$, we have 
	$$
	\delta(\Mcal)\ge {1\over 6}\left({\log n\over n}\right)^{1/d}.
	$$
	Therefore, we can conclude that when $c=1/6$,
	$$
	\PP_0(T=1)+\PP_1(T=0)\to 1.
	$$

	
\subsection{Proof for Theorem~\ref{thm:loweraiml}}

We follow a similar strategy and the same notations in proof for Theorem~\ref{thm:lowergeneral}, but need a different way to construct hypotheses. 

\paragraph{Step 1: hypothesis construction} Same to the proof for Theorem~\ref{thm:lowergeneral}, we assume the observed data is drawn from a uniform distribution on the single manifold $\Mcal_0$ under $\HH_0$. The hypothesis $\HH_1$ is constructed in a different way from the proof for Theorem~\ref{thm:lowergeneral}. Specifically, we  divide the space $[0,1]^{d_s}$ into $L=M^{d_s}$ cubes of size $M^{-1}\times \ldots \times M^{-1}$ for some positive integer $M$, which will be specified later. We still name these disjoint cubes $Q_1,\ldots, Q_L$ and define $\tilde{Q}_l$ as a cube with the same center of $Q_l$ but a smaller size  $(3M)^{-1}\times \ldots \times (3M)^{-1}$. Under $\HH_1$, we consider the following manifolds
$$
\Mcal_1(Q_l)=\{(x,y,0_{D-d}):x\in[0,1]^{d_s}\setminus Q_l, y\in [0,1]^{d_v}\}
$$
and
$$
\Mcal_2(Q_l)=\{(x,y,0_{D-d}):x\in\tilde{Q}_l,  y\in [0,1]^{d_v}\}.
$$
If we define $\Mcal(Q_l)=\Mcal_1(Q_l) \cup \Mcal_2(Q_l)$, we have $\delta(\Mcal(Q_l))=(3M)^{-1}$. Given each pair of $Q_l$ and $\tilde{Q}_l$, we define a uniform distribution over $\Mcal(Q_l)$ as hypothesis $\HH_1(Q_l)$ and write $\PP_1(Q_l)$ as the corresponding probability distribution of observed data $X_1,\ldots,X_n$. $\PP_1$ is defined as a mixture distribution of $\PP_1(Q_l)$
$$
\PP_1={1\over L}\sum_{l=1}^L\PP_1(Q_l).
$$

\paragraph{Step 2: bounding $\chi^2$ divergence} We can bound the $\chi^2$ divergence in the same way as the proof for Theorem~\ref{thm:lowergeneral}. Specifically, we can show
$$
\chi^2(\PP_1,\PP_0)\le {1\over L}\exp \left({n\Delta\over 1-\Delta}\right)+{n\Delta^2\over (1-\Delta)\sqrt{1-2\Delta}},
$$
where $\Delta=M^{-d_s}-(3M)^{-d_s}$.

\paragraph{Step 3: applying theorem of fuzzy hypothesis} We can choose $M=\lceil (2n/\log n)^{1/d_s} \rceil$ and still show 
$$
\PP_0(T=1)+\PP_1(T=0)\to 1.
$$

\subsection{Proof for Theorem~\ref{thm:downstream}}
	Without loss of generality, we can assume $Y_i=1$ when $\tilde{X}_i\in\Mcal_s$ for $1\le s\le K'$ and $Y_i=-1$ if $\tilde{X}_i\in\Mcal_s$ for $K'< s\le K$. Furthermore, we assume the first $K$ eigenfunctions have the following forms
	$$
	\theta_s(x)=\begin{cases}1,& x\in \Mcal_s\\0, & x\in \Mcal\setminus\Mcal_s \end{cases}.
	$$
	
	\paragraph{Step 1} In this step, we aim to show that the learned representation $\hat{\Theta}(\tilde{X}_1),\ldots, \hat{\Theta}(\tilde{X}_m)$ are linearly separable with respect to $Y$ in a large probability. An application of Markov's inequality suggests 
	$$
	\PP\left(|\hat{\theta}_s(\tilde{X}_i)-\theta_s(\tilde{X}_i)|>{1\over 3K}\right)\le {\EE(\hat{\theta}_s(\tilde{X}_i)-\theta_s(\tilde{X}_i))^2\over (1/3K)^2}\le 9K^2\chi_n.
	$$
	Therefore, with probability $1-9mK^3\chi_n$, we have 
	\begin{equation}
		\label{eq:constraint}
		|\hat{\theta}_s(\tilde{X}_i)-\theta_s(\tilde{X}_i)|\le {1\over 3K},\qquad 1\le s\le K,\ 1\le i\le m.
	\end{equation}
	Let $\beta^o=(\beta^o_s)$ such that $\beta^o_s=1$ when $1\le s\le K'$ and $\beta^o_s=-1$ when $K'< s\le K$. When $Y_i=1$, then
	$$
	{\beta^o}^T\hat{\Theta}(\tilde{X}_i)\ge 1-{1\over 3K}-(K'-1){1\over 3K}-(K-K'){1\over 3K}\ge {2\over 3}>0.
	$$
	On the other hand, if $Y_i=-1$, then
	$$
	{\beta^o}^T\hat{\Theta}(\tilde{X}_i)\le K'{1\over 3K}+(K-K'-1){1\over 3K}-\left(1-{1\over 3K}\right)\le -{2\over 3}<0.
	$$
	Therefore, $\{\hat{\Theta}(\tilde{X}_i):Y_i=1\}$ and $\{\hat{\Theta}(\tilde{X}_i):Y_i=-1\}$ are linearly separable. 
	
	\paragraph{Step 2} The goal of this step is to identify a set $\Bcal$ which covers all converged weight vector $\beta$ in the logistic regression. According to \cite{soudry2018implicit,ji2018risk}, the  weight vector $\beta$ of logistic regression converges to the direction of the max-margin solution. Specifically, Theorem 1.1 in \cite{ji2018risk} suggests that 
	$$
	\lim_{t\to\infty}{\beta_t\over \|\beta_t\|}={\beta^\ast\over \|\beta^\ast\|},
	$$
	where $\beta^\ast$ is the optimal solution of the following optimization problem
	\begin{equation}
		\label{eq:hardsvm}
		\min_\beta \|\beta\|^2,\qquad {\rm s.t.}\quad Y_i\beta^T\hat{\Theta}(\tilde{X}_i)\ge 1,\quad 1\le i\le n.
	\end{equation}
	Now we study the property of $\beta^\ast$. Clearly, $3\beta^o/2$ satisfy the constraint in \eqref{eq:hardsvm} and therefore we have 
	$$
	\|\beta^\ast\|^2\le \|3\beta^o/2\|^2=9K/4.
	$$
	When $\tilde{X}_i\in \Mcal_s$ for some $1\le s\le K'$, \eqref{eq:constraint} implies
	$$
	\beta^T\hat{\Theta}(\tilde{X}_i)\le \beta_s+{1\over 3K}\sum_{s=1}^K|\beta_s|\le \beta_s+{1\over 3K}\|\beta\|\sqrt{K}.
	$$
	An application of $\|\beta^\ast\|^2\le 9K/4$ yields
	$$
	1\le {\beta^\ast}^T\hat{\Theta}(\tilde{X}_i)\le \beta^\ast_s+{1\over 3K}\|\beta^\ast\|\sqrt{K}\le \beta^\ast_s+{1\over 2}.
	$$
	Therefore, we can show that $\beta^\ast_s\ge 1/2$ when $1\le s\le K'$. Through the same way, we can show that $\beta^\ast_s\le -1/2$ when $K'< s\le K$. Overall, we can conclude 
	$$
	\beta^\ast\in\Bcal:=\left\{\beta\in \RR^K: \|\beta\|^2\le {9K\over 4},\quad \beta_s\ge {1\over 2}\ {\rm if}\ 1\le s\le K'\quad {\rm and}\quad\beta_s\le -{1\over 2} \ {\rm if}\  K'< s\le K \right\}.
	$$ 
	
	\paragraph{Step 3} 
	Because $\beta^\ast\in\Bcal$, it is sufficient to study the misclassification rate of the following classifier
	$$
	\hat{H}_{\hat{\Theta},\beta}(x)=\begin{cases}
		1,& \beta^T\hat{\Theta}(x)>0\\
		-1,& \beta^T\hat{\Theta}(x)\le0
	\end{cases}
	$$
	for any $\beta\in \Bcal$. Recall the true label is 
	$$
	H^\ast(x)=\begin{cases}
		1,& x\in \Mcal_s\ {\rm when}\ 1\le s\le K'\\
		-1,& x\in \Mcal_s \ {\rm when}\ K'< s\le K
	\end{cases}.
	$$
	When $\beta\in \Bcal$, we can have
	\begin{align*}
		\PP(\hat{H}_{\hat{\Theta},\beta}(X)\ne H^\ast(X))\le \PP\left(|\hat{\theta}_s(X)-\theta_s(X)|>{1\over 6K},\ 1\le s\le K\right)
	\end{align*}
	because when $|\hat{\theta}_s(X)-\theta_s(X)|\le{1/6K}$, 
	$$
	\beta^T\hat{\Theta}(X)\ge \beta_s-{1\over 6K}\sum_s|\beta_s|\ge {1\over 2}-{1\over 6K}{3K\over 2}\ge {1\over 4}\qquad {\rm if }\ X\in \Mcal_s\ {\rm for \ some\ }1\le s\le K'
	$$
	and 
	$$
	\beta^T\hat{\Theta}(X)\le \beta_s+{1\over 6K}\sum_s|\beta_s|\le -{1\over 2}+{1\over 6K}{3K\over 2}\le -{1\over 4}\qquad {\rm if }\ X\in \Mcal_s\ {\rm for \ some\ }K'< s\le K.
	$$
	By union bound and Markov's inequality, we have
	\begin{align*}
		\PP\left(|\hat{\theta}_s(X)-\theta_s(X)|>{1\over 6K},\ 1\le s\le K\right)&\le \sum_{s=1}^K\PP\left(|\hat{\theta}_s(X)-\theta_s(X)|>{1\over 6K}\right)\\
		&\le K{\EE(\hat{\theta}_s(X)-\theta_s(X))^2\over (1/6K)^2}\\
		&\le 36K^3\chi_n.
	\end{align*}
	Therefore, we can conclude that with probability $1-9mK^3\chi_n$,
	$$
	\xi(\hat{\Theta})\le 36K^3\chi_n.
	$$

\subsection{Proof for Proposition~\ref{prop:cross}}
	Let $k(X_i)$ be the manifold that $X_i$ belongs to, i.e., $k(X_i)=k$ if $X_i\in \Mcal_k$. The following analysis is conducted by conditioning on $k(X_1),\ldots,k(X_n)$. We write the number of edges connecting points in $\Mcal_{k_1}$ and $\Mcal_{k_2}$ as
	$$
	Z_{k_1,k_2}=\sum_{X_i\in \Mcal_{k_1}, X_j\in \Mcal_{k_2}}\bI(\|X_i-X_j\|\le r).
	$$
	The expectation of $Z_{k_1,k_2}$ can be written as
	\begin{align*}
		\EE(Z_{k_1,k_2})&=\EE\left(\sum_{X_i\in \Mcal_{k_1}, X_j\in \Mcal_{k_2}}\bI(\|X_i-X_j\|\le r)\middle| k(X_1),\ldots,k(X_n)\right)\\
		&\le n_{k_1}\sup_{x\in \Mcal_{k_1}}\EE\left(\sum_{X_j\in \Mcal_{k_2}}\bI(\|x-X_j\|\le r)\middle| k(X_1),\ldots,k(X_n) \right) \\
		&\le \beta n_{k_1}n_{k_2} 
	\end{align*}
	To bound the variance of $Z_{k_1,k_2}$, we apply Efron-Stein inequality \citep[See Theorem 3.1 in][]{boucheron2013concentration}
	\begin{align*}
		\Var(Z_{k_1,k_2})\le & {1\over 2}\sum_{i=1}^{k_1} \EE\left(\sum_{X_j\in \Mcal_{k_2}}\bI(\|X_i-X_j\|\le r)-\bI(\|X'_i-X_j\|\le r)\right)^2\\
		&+{1\over 2}\sum_{j=1}^{k_2} \EE\left(\sum_{X_i\in \Mcal_{k_1}}\bI(\|X_i-X_j\|\le r)-\bI(\|X_i-X'_j\|\le r)\right)^2,
	\end{align*}
	where $X'_i$ and $X'_j$ are independent copies of $X_i$ and $X_j$. The first term of the right hand side in above inequality can be written as
	\begin{align*}
		&\EE\left(\sum_{X_j\in \Mcal_{k_2}}\bI(\|X_i-X_j\|\le r)-\bI(\|X'_i-X_j\|\le r)\right)^2\\		
		\le & 4\Var\left(\sum_{X_j\in \Mcal_{k_2}}\bI(\|X_i-X_j\|\le r)\right)\\
		\le & 4\sup_{x\in \Mcal_{k_1}}\Var\left(\sum_{X_j\in \Mcal_{k_2}}\bI(\|x-X_j\|\le r)\right)+4\EE\left( \EE\left(\sum_{X_j\in \Mcal_{k_2}}\bI(\|X_i-X_j\|\le r)\middle|X_i\right) \right)^2\\
		\le & 4\sup_{x\in \Mcal_{k_1}}\left(\sum_{X_j\in \Mcal_{k_2}}\Var\left(\bI(\|x-X_j\|\le r)\right)\right)+4\beta^2n_{k_2}^2\\
		\le & 4\beta n_{k_2}+4\beta^2n_{k_2}^2.
	\end{align*}
	Here we use the fact
	$$
	0\le \EE\left(\sum_{X_j\in \Mcal_{k_2}}\bI(\|X_i-X_j\|\le r)\middle|X_i\right) \le \beta n_{k_2}.
	$$
	We can apply the same strategy to bound the other term of the right hand side to obtain 
	$$
	\EE\left(\sum_{X_i\in \Mcal_{k_1}}\bI(\|X_i-X_j\|\le r)-\bI(\|X_i-X'_j\|\le r)\right)^2\le 4\beta n_{k_1}+4\beta^2n_{k_1}^2.
	$$
	Putting two terms together yields
	$$
	\Var(Z_{k_1,k_2})\le 4\beta n_{k_1}n_{k_2}+2\beta^2n_{k_1}n_{k_2}(n_{k_1}+n_{k_2}).
	$$ 
	By Chebyshev's inequality, we can conclude that
	$$
	\PP\left(Z_{k_1,k_2}\ge 2\left(\beta n_{k_1}n_{k_2}+t\sqrt{\beta n_{k_1}n_{k_2}}+t\beta\sqrt{n_{k_1}n_{k_2}(n_{k_1}+n_{k_2})} \right)\right)\le {1\over t^2}.
	$$
	By union bound and the fact 
	$$
	\PP(nw_k/2 \le n_{k}\le 2nw_k,\forall 1\le k\le K)\ge 1-2K\exp(-cn),
	$$
	we have 
	$$
	\PP\left(\max_{1\le k_1,k_2\le K}Z_{k_1,k_2}\ge c\beta n^2+t\left(\sqrt{\beta}n+\beta n^{3/2}\right)\right)\le 2K\exp(-cn)+{K^2\over t^2}
	$$
	If we apply Proposition A.4 in \cite{trillos2021large}, we can conclude that 
	$$
	b_C(\tilde{P}\theta)\le {C\left(\beta n +t\sqrt{\beta}+t\beta\sqrt{n}\over nr^{d+2}\right)}\left(1+\lambda^{d/2+2}\right)\|\theta\|^2_{L^2(\pi)}.
	$$

\end{appendices}

\end{document}